\documentclass[10pt,twocolumn,letterpaper]{article}

\usepackage[pagenumbers]{cvpr} 

\newcommand{\red}[1]{{\color{red}#1}}

\definecolor{cvprblue}{rgb}{0.21,0.49,0.74}
\definecolor{citecolor}{HTML}{0071bc}
\usepackage[pagebackref,breaklinks=true,letterpaper=true,colorlinks,citecolor=citecolor,bookmarks=false]{hyperref}

\usepackage{multirow}
\usepackage{multicol}
\usepackage{colortbl}
\usepackage{amsmath}
\usepackage{amssymb}
\usepackage{booktabs}
\usepackage{dsfont}
\usepackage{soul, nicefrac}
\usepackage{graphicx}
\graphicspath{{./images/}}
\usepackage{wrapfig}
\usepackage{changepage}
\usepackage{sidecap}
\usepackage{tcolorbox}

\usepackage{booktabs}
\usepackage{mathrsfs} 
\usepackage{eucal}
\usepackage{listings}
\usepackage{comment}
\usepackage{pifont}
\usepackage{xcolor}
\usepackage{enumitem}
\usepackage{tikz}
\usepackage{mdframed}
\usepackage{etoolbox}
\newcommand{\circled}[2][]{\tikz[baseline=(char.base)]
    {\node[shape = circle, draw, inner sep = 1pt]
    (char) {\phantom{\ifblank{#1}{#2}{#1}}};%
    \node at (char.center) {\makebox[0pt][c]{#2}};}}
\robustify{\circled}

\usepackage{caption}
\def\prob{{\mathbf{P}}}
\DeclareMathOperator*{\argmax}{arg\,max}
\def\f{{\mathbf{f}}}
\def\x{{\mathbf{x}}}

\usepackage[capitalize]{cleveref}
\crefname{section}{Sec.}{Secs.}
\Crefname{section}{Section}{Sections}
\Crefname{table}{Table}{Tables}
\crefname{table}{Table}{Tables}

\definecolor{mygray}{RGB}{230,230,230}
\definecolor{headcolor}{HTML}{4472c4}
\definecolor{tailcolor}{HTML}{ed7c31}
\definecolor{extracolor}{HTML}{bf9000}
\definecolor{darkgreen}{HTML}{228B22}

\newcommand{\head}[1]{\textcolor{headcolor}{#1}}
\newcommand{\tail}[1]{\textcolor{tailcolor}{#1}}
\newcommand{\aux}[1]{\textcolor{extracolor}{#1}}
\usepackage{xcolor} 

\def\paperID{3247} 
\def\confName{CVPR}
\def\confYear{2024}

\usepackage[normalem]{ulem}

\let\originalleft\left
\let\originalright\right
\renewcommand{\left}{\mathopen{}\mathclose\bgroup\originalleft}
\renewcommand{\right}{\aftergroup\egroup\originalright}

\newcommand{\tablestylesmaller}[2]{\setlength{\tabcolsep}{#1}\renewcommand{\arraystretch}{#2}\centering\footnotesize}

\def\paperID{2983} 
\def\confName{CVPR}
\def\confYear{2025}

\title{Learning from Neighbors: Category Extrapolation for Long-Tail Learning}

\author{
Shizhen Zhao\textsuperscript{\rm 1} 
\;\; Xin Wen\textsuperscript{\rm 1} 
\;\; Jiahui Liu\textsuperscript{\rm 1} 
\;\; Chuofan Ma\textsuperscript{\rm 1} 
\;\; Chunfeng Yuan\textsuperscript{\rm 2} 
\;\; Xiaojuan Qi\textsuperscript{\rm 1 \thanks{Corresponding author}}
\\
\;\textsuperscript{\rm 1} The University of Hong Kong \\ \ \textsuperscript{\rm 2} National Laboratory of Pattern Recognition, Institute of Automation, Chinese Academy of Sciences \\
\ \texttt{\{zhaosz,xjqi\}@eee.hku.hk}
}

\begin{document}
\maketitle

\begin{abstract}
Balancing training on long-tail data distributions remains a long-standing challenge in deep learning. While methods such as re-weighting and re-sampling help alleviate the imbalance issue, limited sample diversity continues to hinder models from learning robust and generalizable feature representations, particularly for tail classes.
In contrast to existing methods, we offer a novel perspective on long-tail learning, inspired by an observation: datasets with finer granularity tend to be less affected by data imbalance. In this paper, we investigate this phenomenon through both quantitative and qualitative studies, showing that increased granularity enhances the generalization of learned features in tail categories. 
Motivated by these findings, we propose a method to increase dataset granularity through category extrapolation. Specifically, we introduce open-set fine-grained classes that are related to existing ones, aiming to enhance representation learning for both head and tail classes. 
To automate the curation of auxiliary data, we leverage large language models (LLMs) as knowledge bases to search for auxiliary categories and retrieve relevant images through web crawling. To prevent the overwhelming presence of auxiliary classes from disrupting training, we introduce a neighbor-silencing loss that encourages the model to focus on class discrimination within the target dataset. 
During inference, the classifier weights for auxiliary categories are masked out, leaving only the target class weights for use.  Extensive experiments on three standard long-tail benchmarks demonstrate the effectiveness of our approach, notably outperforming strong baseline methods that use the same amount of data. The code will be made publicly available. 
\end{abstract}

\section{Introduction}

Deep models have shown extraordinary performance on large-scale curated datasets~\citep{he2016deep,karen2015vggnet,dosovitskiy2021vit,liu2025can}.
But when dealing with real-world applications, they generally face highly imbalanced (\eg, long-tailed) data distribution: instances are dominated by a few head classes, and most classes only possess a few images~\citep{wang2021ride,he2021dive,LFME,dong2023lpt}.
Learning in such an imbalanced setting is challenging as the instance-rich (or head) classes dominate the training procedure~\citep{cui2021paco,samuel2021dro,LTR-WD,zhong2021mislas}.
Without considering this situation, models tend to classify tailed class samples as similar head categories, leading to significant performance degradation on tail categories~\citep{IDR,CMO,CKT,zhu2022balanced}. 

\begin{figure}[t]
\centering
\includegraphics[width=1\linewidth]{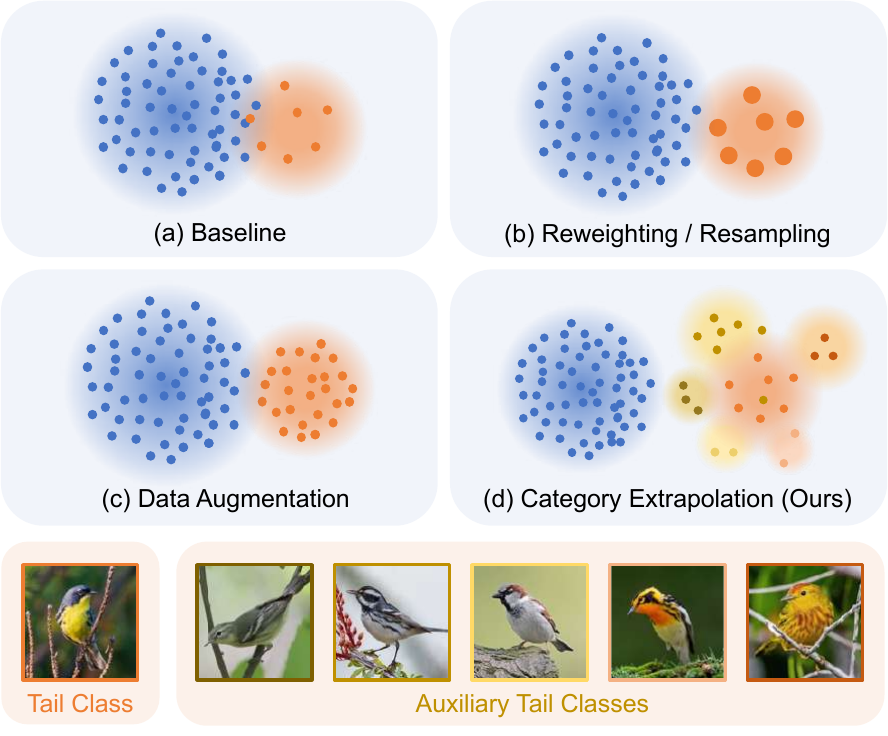}
\caption{{\bf Holistic comparison to previous philosophy.} (a) Data imbalance between \head{head} and \tail{tail} classes makes biased features; (b, c): Previous methods are still bounded by existing known classes; (d) We instead seek help from \aux{auxiliary} open-set data.} \label{fig:method-compare}
\end{figure}

Existing works tackle challenges in long-tail learning from various perspectives.
An earlier stream is to re-balance the learning signal (\eg, re-weighting~\citep{cui2019class} and re-sampling~\citep{chawla2002smote}). 
Yet, they inevitably face the scarcity of data and suffer from over-fitting on tail classes (\cref{fig:method-compare}\red{b}). 
Another straightforward fix is to augment training samples into diverse ones through image transformations~\citep{ref:cutout_2017,ref:mixup_iclr2018,ref:cutmix_iccv2019,chou2020remix}. 
These methods typically increase the loss weights or enhance the sample diversity of tail classes to balance representation learning (\cref{fig:method-compare}\red{c}). 
Despite advances, limited sample diversity still constrains the ability to generalize the learned features. Additionally, improvements in tail class performance are often accompanied by a decline in head class performance. 
This limitation motivates us to investigate what factors contribute to generalizable feature learning in long-tail settings. Our insight is inspired by a common, yet counterintuitive, phenomenon observed in existing benchmarks: despite being more imbalanced than ImageNet-LT~\citep{liu2019oltr}, iNat18~\citep{van2018inat} achieves nearly balanced performance (see \cref{tab:avg_perf}). 
This observation raises the question: \textit{Does granularity play a role in the performance balance of long-tail learning?}

\begin{table}[h]
\centering
\tablestylesmaller{3.3pt}{1}
\begin{tabular}{lllllccc}
\toprule
Dataset  & \#Class & \#Train & Granul.  & Imb. Factor $\beta$ &  Many  &  Med.  &  Few \\
\cmidrule(r){1-5} \cmidrule{6-8}
IN-LT    &  1000   &  116K   & Coarse  & 1280/5=256     & 68.2   & 56.8     & 41.6  \\
iNat18   &  8142   &  438K   & Fine    & 1000/2=500     & 70.3   & 71.3     & 70.2  \\
\bottomrule
\end{tabular}
\vspace{.0em}
\caption{Average performance of previous methods.} \label{tab:avg_perf}
\vspace{-1em}
\end{table}

\begin{figure*}[t]

\begin{center}
    \begin{subfigure}{0.24\textwidth}
        \includegraphics[width=\textwidth]{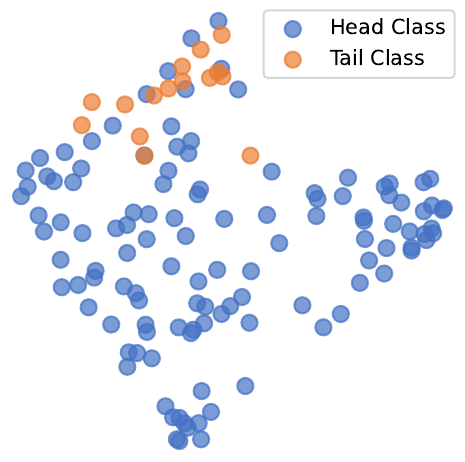}
        \caption{Raw feature space (train).}
        \label{fig:feature_vis_a}
    \end{subfigure}
    \hfill
    \begin{subfigure}{0.24\textwidth}
        \includegraphics[width=\textwidth]{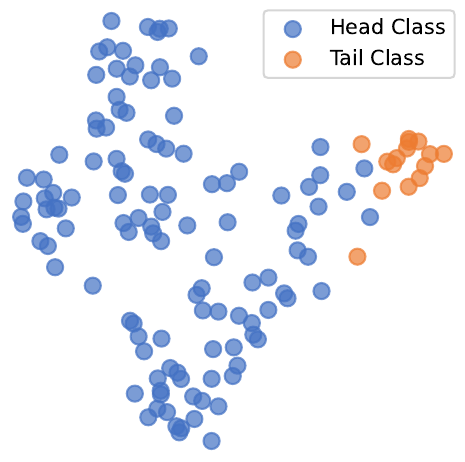}
        \caption{Baseline after training (train).}
        \label{fig:feature_vis_b}
    \end{subfigure}
    \hfill
    \begin{subfigure}{0.24\textwidth}
        \includegraphics[width=\textwidth]{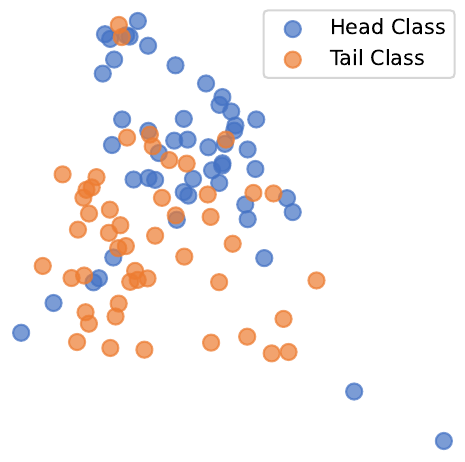}
        \caption{Baseline after training (val).}
        \label{fig:feature_vis_c}
    \end{subfigure}
    \hfill
    \begin{subfigure}{0.24\textwidth}
        \includegraphics[width=\textwidth]{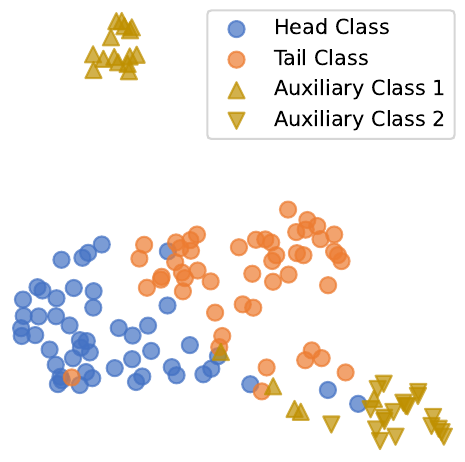}
        \caption{After training w/ aux. class (val).}
        \label{fig:feature_vis_d}
    \end{subfigure}
    \hfill
   \caption{\textbf{Feature visualization of confusing \head{head} and \tail{tail} classes by UMAP~\citep{mcinnes2020umap} on ImageNet-LT~\citep{liu2019oltr}.} (a) Raw feature space of training data by DINOv2~\citep{oquab2023dinov2}; (b) Feature space of training data after the training phase; (c) The baseline (re-weighting) shows poor generalization on validation data; (d) Adding auxiliary categories condenses clusters and improves separation.} \label{fig:feature_vis}
\end{center}
\vspace{-0.7cm}
\end{figure*}

To investigate this further, we conducted a pilot study (see \cref{subsec:granularity_effect}) using a larger data pool and controlled experiments to verify this phenomenon. We found that datasets with finer granularity are less affected by data imbalance.
Feature visualizations (see \cref{fig:feature_vis}) reveal that, despite a long-tail distribution, datasets with finer granularity enable the model to learn more generalized representations. This discovery motivates us to explore \textit{altering data distribution by introducing open-set categories to increase the granularity of data for long-tail learning.}


At the core of our approach is the idea of augmenting training data with fine-grained categories related to the original ones, thereby increasing granularity (\cref{fig:method-compare}\red{d}). 
To acquire auxiliary data, we establish a fully automated data crawling pipeline powered by the knowledge of large language models (LLMs). 
Specifically, for each class to be expanded, we query an LLM for $k$ fine-grained auxiliary classes, then retrieve corresponding images from the web based on these class names (\cref{fig:method-search}). 
The crawled data are subsequently integrated with the original dataset for model training.   
During training, we introduce a neighbor-silencing loss to enhance discrimination between confusing classes, prevent the model from being overwhelmed by auxiliary classes, and ensure alignment with the objectives of the testing phase. After training, the classifier by simply masking out the auxiliary classes demonstrates strong performance without the need for additional classifier re-balancing, as required in previous methods~\citep{kang2020decoupling,zhou2020bbn}.

Intuitively, our method could be interpreted as \textit{category extrapolation}. These augmented categories complete the learning signal, which may fill the gap between originally distinct classes, encourage continuity and smoothness of the feature manifold, and allow better generalization of representations across classes. In terms of classification, samples of auxiliary classes take up the neighborhood of existing classes, thus explicitly enlarging the margin between them and encouraging discriminability. Empirically, we indeed observe tighter clusters and better separation in-between (\cref{fig:feature_vis_d}).

Our major contributions are summarized as follows: 
\begin{itemize}
\item 
We explore the effect of granularity on the performance balance in long-tail learning, which motivate us to introduce neighbor classes to increase the granularity and facilitate representation learning for both head and tail classes. 
\item We propose a neighbor-silencing learning loss to facilitate long-tail learning with extra open-set categories and design a fully automatic data acquisition pipeline to efficiently harvest data from the Web. 
\item We conduct extensive experiments across standard benchmarks using various training paradigms (\eg, random initialization, CLIP~\citep{radford2021clip}, and DINOv2~\citep{oquab2023dinov2}), all of which consistently demonstrate high performance. 

\end{itemize} 

\section{Pilot Study} \label{sec:study}
In this section, we investigate whether granularity impacts performance balance in long-tail distribution. 
We first provide preliminary for long-tail learning and an analysis on a baseline method in \cref{subsec:prelim}. 
Then, we verify the impact of the granularity of training data on long-tail learning (\cref{subsec:granularity_effect}) from both quantitative and qualitative perspectives. 
%



\subsection{Preliminary} \label{subsec:prelim}
In long-tail visual recognition, the model has access to a set of $N$ training samples $\mathcal{S} = \{(\x_n, y_n)\}^N_{n=1}$, where $\x_n \in \mathcal{X} \subset \mathbb{R}^D$ and labels $\mathcal{Y}=\{1,2,..,L\}$. Training class frequencies are defined as $N_y = \sum_{(x_n, y_n) \in\mathcal{S}} \mathds{1}_{y_n=y}$  and the test-class distribution is assumed to be sampled from a uniform distribution over $\mathcal{Y}$
, but is not explicitly provided during training. A classic solution is to minimize the balanced error (BE), of a scorer $\f:\mathcal{X}\rightarrow\mathbb{R}^L$, defined as:
\begin{equation} \label{eq:be}
    {\rm BE}\left(\x,\f(\cdot)\right) = \sum_{y\in\mathcal{Y}}  \prob_{\x|y} \left(y \notin \argmax_{y' \in \mathcal{Y}} \f_{y'}(\x)\right) ,
\end{equation}
where $\f_y(x)$ is the logit produced for true label $y$ for sample $\x$. Traditionally, this is done by minimizing a proxy loss, the Balanced Softmax Cross Entropy (BalCE)~\citep{cui2019class}:
\begin{equation} \label{eq:bal-ce}
    \begin{aligned}
        &\mathcal{L}_{\text{BalCE}}\left(\mathcal{M}(\mathbf{x}|\theta_f,\theta_w), \mathbf{y}_i\right) = - \log\left[p(\mathbf{y}_i|\mathbf{x};\theta_f,\theta_w)\right] \\
        &= - \log\left[\frac{n_{\mathbf{y}_i} e^{z_{\mathbf{y}_i}}}{\sum_{\mathbf{y}_j \in \mathcal{Y}} n_{\mathbf{y}_j} e^{z_{\mathbf{y}_j}}}\right] \\
        &= \log\left[1+\sum_{\mathbf{y}_j\neq \mathbf{y}_i} e^{\log n_{\mathbf{y}_j} - \log n_{\mathbf{y}_i} + \mathbf{z}_{\mathbf{y}_j} - \mathbf{z}_{\mathbf{y}_i}}\right].
    \end{aligned}
\end{equation}
This is known as \textit{re-weighting}, where the contribution of each label's individual loss is scaled by an inverse class frequency derived from the class's instance number $n_{y_i}$. We adopt this setting as the baseline in follow-up experiments.

\noindent \textbf{On the failure of re-balancing.}
The primary challenge of long-tail learning stems from data imbalance, which affects the representation learning of both head classes and few-shot classes. 
For head classes, if there is a lack of effective negative-class samples, then learning an effective boundary is challenging. 
To better demonstrate this, we provide feature visualizations of confusing head (\head{Scottish Deerhound}) and tail (\tail{Irish Wolfhound}) classes on ImageNet-LT in \cref{fig:feature_vis}.
As in \cref{fig:feature_vis_a}, these two classes are challenging even for the advanced vision foundation model DINOv2~\citep{oquab2023dinov2}.
After training with the re-weighting baseline on the imbalanced training data, the learned features seem relatively satisfactory (\cref{fig:feature_vis_b}). However, the generalization is poor: samples in the validation data are still convoluted, and the separation between them is unclear (\cref{fig:feature_vis_c}).
On top of this baseline, we then study the effect of data distribution on long-tail learning. 

\begin{figure}[]
\centering
\includegraphics[width=0.9\linewidth]{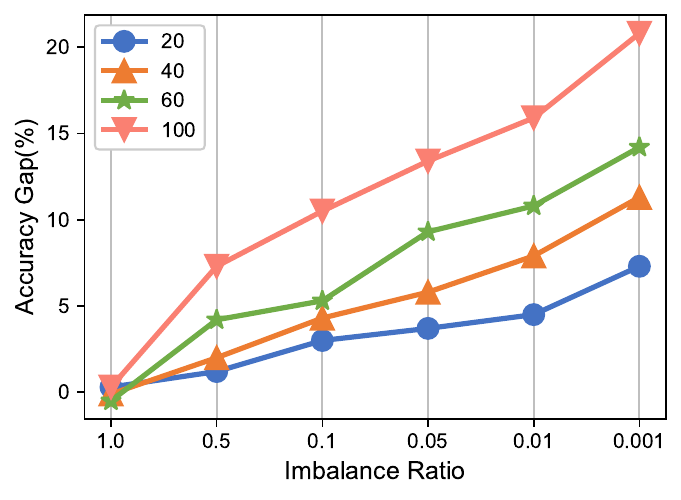}
\vspace{-0.1cm}
\caption{Effect of granularity \vs imbalance ratio.} \label{fig:factor_granularity}
\vspace{-0.4cm}
\end{figure}

\subsection{Granularity Matters in Long-Tail learning} \label{subsec:granularity_effect}


We study whether the granularity of the dataset is critical to long-tail learning. 
Our study is motivated by an intriguing observation that, although more classes and stronger imbalance, we observe nearly balanced performance on iNat18~\citep{van2018inat}, as opposed to ImageNet-LT~\citep{liu2019oltr}. 
A significant distinction is that iNat18 is an extremely fine-grained dataset with over 8000 categories, yet it only consists of 14 superclasses in total. 
On the other hand, ImageNet-LT, although comprising only 1000 categories, has over 100 superclasses, making it relatively coarse-grained.  
Therefore, we conduct experiments to study the effect of granularity on long-tail learning.


\textit{Dataset Configuration.} To this end, we construct a dataset pool using ImageNet-21k~\citep{ridnik2021imagenet21k} and OpenImage~\citep{openimages} datasets. To investigate the influence of granularity, we sample 500 classes from the pool for each time and control the number of superclasses to be $\{20, 40, 60, 100\}$  based on WordNet. 
Then, we used different imbalance ratios $\{1.0, 0.5, 0.1, 0.05, 0.01, 0.001\}$ to study the effect of granularity on the imbalance ratio. 
We train the model (ViT-Base~\citep{dosovitskiy2021vit})  using BalCE~\citep{cui2019class} as \cref{eq:bal-ce}.
We conduct 5 experiments and take the average value.

%
In \cref{fig:factor_granularity}, we show the performance gap between head categories and tail categories under different dataset imbalance ratios.
The results show that as the granularity increases, the dataset is less sensitive to the imbalance ratio. 
For example, when the number of superclasses is 20, the performance gap between the head and tail is 7.3\%, while the gap is 20.8\% when the number of superclasses is 100, under the severe imbalance (imbalance ratio=0.001). 

\begin{mdframed}[backgroundcolor=gray!15] 
\noindent\textbf{Finding 1:}  Increased granularity of training data benefits long-tail learning. 
\end{mdframed}

In a fine-grained long-tail dataset, although there are few samples for tail categories, many categories share similar patterns, which is conducive to learning distinctive features, thus enhancing generalizability. 
As reflected in \cref{fig:feature_vis_d}, for clearer visualization, we sample two fine-grained categories that is denoted as the auxiliary classes. 
The visualization shows that the separation between head and tail classes is clearly improved. 
Also, the distribution of intra-class samples is also more compact. 
Due to the space limitation, we show more examples in Appendix. 
This motivate us to introduce diverse open-set auxiliary categories to enhance the granularity for close-set long-tail learning. 

\begin{mdframed}[backgroundcolor=gray!15] 
\noindent\textbf{Finding 2:} Despite long-tail distribution, increased granularity could explicitly separate and condensify existing data clusters.
\end{mdframed}

Based on the above findings, given a long-tail dataset, we aim to establish a framework that can effectively acquire auxiliary data to enhance the granularity.   
Specifically, we utilize LLMs to query the candidate auxiliary categories and crawl images from the Web, followed by a filtering stage to ensure similarity and diversity. 
To better incorporate auxiliary data for training with target categories, we propose a 
Neighbor-Silencing Loss to avoid being overwhelmed by auxiliary classes. 
Details are included in \cref{sec:method}.

\section{Long-Tail Learning by Category Extrapolation}
\label{sec:method}
In this section, we first introduce our simple and automatic pipeline for obtaining auxiliary data in \cref{subsec:data_searching}. Then, we present our new learning objective that effectively leverages the auxiliary data to enhance long-tail learning in \cref{subsec:training_method}.


\subsection{Neighbor Category Searching} \label{subsec:data_searching}
\begin{figure}[t]
\centering
\includegraphics[width=\linewidth]{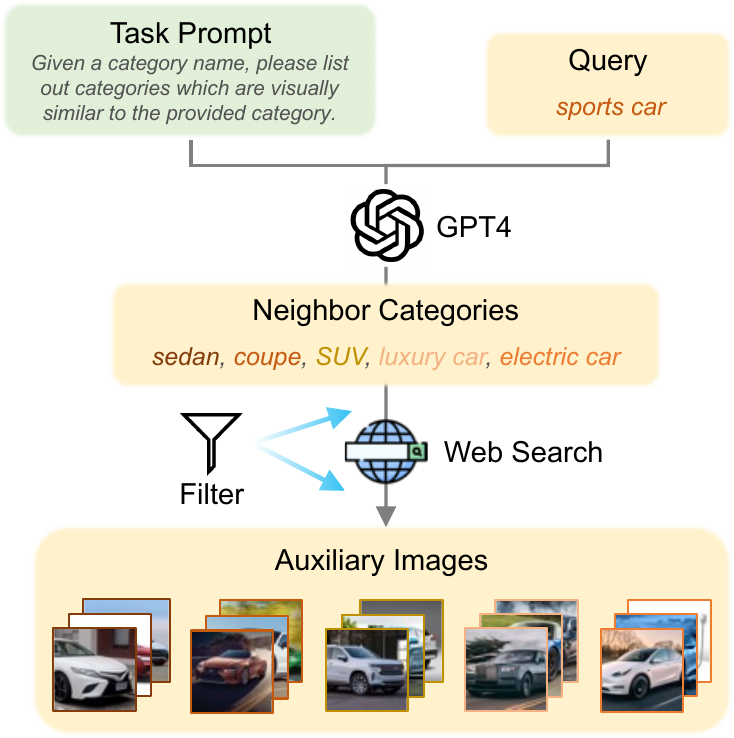}
\caption{{\bf Data crawling pipeline.} We prompt GPT-4~\cite{openai2023gpt4} for fine-grained categories related to query classes and retrieve corresponding images from the web. Classes already in the label set and images of lower visual similarity than the threshold are filtered out.} \label{fig:method-search}
\vspace{-0.3cm}
\end{figure}

In search of neighbor categories sharing some common visual patterns with the pre-defined categories in the dataset, we design a fully automatic crawling pipeline that includes (i) querying neighbor categories from LLMs to obtain similar categories and enhance the granularity of the training data and (ii) retrieving corresponding images from the web and conducting filtering to guarantee similarity and diversity.
An overview of this pipeline is illustrated in \cref{fig:method-search}, and we introduce each step in detail as follows.

\vspace{0.1in}\noindent\textbf{Querying LLM for Neighbor Categories.}
We take advantage of the recent development of Large Language Models (LLMs), \eg, GPT-4~\citep{openai2023gpt4}, and query them for expert knowledge of possible fine-grained classes with respect to the classes to extrapolate (\ie, the medium and tail classes by default).
For example, we can prompt the language model with: ``Please create a list which contains 5 fine-grained categories related to {\tt{\{CLS\}}}''.
However, the output of this naive prompt is unstable, possibly because `fine-grained categories' by itself is quite a broad and vague concept. To make the prompt more concrete and clear for LLMs, we design a structural prompt with in-context learning: 

\noindent

\begin{minipage}{0.45\textwidth}
\vspace{1mm}
\begin{tcolorbox} 
\small
{
\noindent\textbf{Task:} Given a category name, please list out 5 classes that are fine-grained categories related to the provided classes.

\vspace{.25em}

\noindent\textbf{Query:} sports car

\vspace{.25em}

\noindent\textbf{Response:} sedan, coupe, SUV, luxury car, electric car

\vspace{.25em}

\noindent\textbf{Query:} {\tt\{CLS\}}

\vspace{.25em}

\noindent\textbf{Response:}
}
\end{tcolorbox}
\vspace{1mm}
\end{minipage}

The LLM then completes the response above. After that, classes in the target dataset $\mathcal{S}$ are filtered out to avoid possible information leaks.
Then, the remaining class names are fed to an image-searching engine for image retrieval.

\vspace{0.1in}\noindent\textbf{Retrieving  and Filtering Images from the Web.}
Images retrieved by the search engine can be noisy, thus, a filtering strategy is adopted.
An image $\mathbf{x}_r$ corresponding to a specific class $y_i$ is dropped if: (i) the class's name does not exist in the associated caption; or (ii) the visual similarity between the class and this image satisfies thresholds: $  \gamma_1 <  {\rm cos}(\mathbf{p}_i, \mathbf{f}_r) < \gamma_2$.
We employ DINOv2~\citep{oquab2023dinov2} for feature extraction and use cosine similarity as the metric.
Specifically, the prototype $\mathbf{p}_i$ of category $y_i$ is computed as the average feature of all samples of this category in the target dataset $\mathcal{S}$: $\mathbf{p}_i=\nicefrac{1}{n_{y_i}}\sum_j\mathbf{f}_j$.
After the filtering process, the model has access to a set of $M$ auxiliary training samples $\mathcal{A} = \{(\x_m, y_m)\}^M_{m=1}$, where $\x_m \in \mathcal{X} \subset \mathbb{R}^D$ and labels $\mathcal{Y}_a=\{L+1,L+2,..,L+K\}$ and the category number for auxiliary set is $K$.

\subsection{Learning with Auxiliary Categories} \label{subsec:training_method}
We mix the auxiliary dataset $\mathcal{A}$ and the target dataset $\mathcal{S}$ for training.
A naive approach is to directly employ the BalCE loss~\citep{cui2019class} by merging the label space:
\begin{equation} \label{eq:aux-ce}
\begin{aligned}
\mathcal{L}_{\text{BalCE}} = - \log\Biggl[n_{\mathbf{y}_i} e^{z_{\mathbf{y}_i}} / ( \overbrace{\sum_{\mathbf{y}_j \in \mathcal{Y}} n_{\mathbf{y}_j} e^{z_{\mathbf{y}_j}}}^{\text{\footnotesize{Target}}}
+ \overbrace{\sum_{\mathbf{y}_j \in \mathcal{Y}_a} n_{\mathbf{y}_j} e^{z_{\mathbf{y}_j}}}^{\text{\footnotesize{Auxiliary}}})\Biggr].
\end{aligned}
\end{equation}
But note that our objective is to classify $L$ categories within the target dataset, as opposed to $L+K$ categories. 
Directly employing the standard BalCE loss as \cref{eq:aux-ce} would result in an inconsistency between the optimization process and the ultimate goal.
The auxiliary part could overwhelm optimization and result in degenerated performance. We thus ``silent'' them by weighting as follows.

\vspace{0.1in}\noindent\textbf{Silencing the Overwhelming Neighbors.}
%
%
Concretely, if $y_j$ is a neighbor category of $y_i$ from auxiliary categories, we spot this as possible neighbor overwhelming and give the corresponding logit a smaller weight.
To clarify,  $y_j$ is a neighbor category of $y_i$ means that $y_j$ is queried from $y_i$ by Neighbor Category Searching (\cref{subsec:data_searching}).
We thus expect the auxiliary classes to influence less the target class which they are queried from, and contribute more to their classification as a whole with respect to other classes. The neighbor-silencing variant of the re-balancing loss is then formulated as:
\begin{equation} \label{eq:sil-ce}
\mathcal{L}_{\text{NS-CE}}=
\log \left[1+\sum_{\mathbf{y}_j\neq \mathbf{y}_i} \lambda_{ij} \cdot e^{\log n_{\mathbf{y}_j} - \log n_{\mathbf{y}_i} + \mathbf{z}_{\mathbf{y}_j} - \mathbf{z}_{\mathbf{y}_i}}\right],
\end{equation}
where $\lambda_{ij}=\lambda_s$, if $y_i$ and $y_j$ satisfy that one is the other's neighbor category and one of them from auxiliary categories, and $\lambda_{ij}=1$ otherwise. $\lambda_s$ is the weight for balancing the loss between neighbor category pairs and non-pairs. By default, $0<\lambda_s<1$.
In this way, we assign a smaller weight to neighbor category pairs, thus, the effect within neighbor classes is weakened, and the optimization focuses more on their separation as a whole from other confusing classes.
%

\vspace{0.1in}\noindent\textbf{Obtaining the Final Classifier.}
Given that our model's classifier includes more categories, it cannot be directly applied to the target dataset for evaluation. 
A common practice is to discard the trained classifier and re-train it with re-balancing techniques on the target dataset through linear probing~\citep{kang2020decoupling,zhou2020bbn}.
However, this could be suboptimal since the separation hyper-planes shaped by auxiliary categories can be undermined. 
Therefore, we try directly masking out the weights of auxiliary categories, retaining only the weights of the target categories.
Specifically, we denote the trained classifier weights as $\mathcal{W} = \{\mathbf{w}_i\}^{L+K}_{i=1}$, where $\mathbf{w}_i \subset \mathbb{R}^C$, and keep $\mathcal{W} = \{\mathbf{w}_i\}^{L}_{i=1}$.
Surprisingly, this simpler approach works better.
This is potentially because incorporating more auxiliary fine-grained categories can enable the classifier to focus on class-specific discriminative features. 
These features possess stronger generalizability, facilitating the classifier to construct more precise separation hyper-planes.

\section{Experiments} \label{sec:exp}



\subsection{Datasets} \label{subsec:dataset}
We experiment with three standard long-tailed image classification benchmarks.
We report accuracy on three splits of the set of classes: Many-shot (more than 100 images), Medium-shot (20$\sim$100 images), and Few-shot (less than 20 images).
Besides, we also report the commonly used top-1 accuracy over all classes for evaluation.
%

\noindent\textbf{ImageNet-LT}~\citep{liu2019oltr} is a class-imbalanced subset of the popular image classification benchmark ImageNet ILSVRC 2012~\citep{Imagenet}. The images are sampled following the \textit{Pareto} distribution with a power value $\alpha=6$, containing 115.8k images from 1,000 categories.
\noindent\textbf{iNaturalist 2018}~\citep{van2018inat} (iNat18 for short) is a species classification dataset, which consists of 437.5k images from 8,142 fine-grained categories following an extreme long-tail distribution.
\noindent\textbf{Places-LT} is a synthetic long-tail variant of the large-scale scene classification dataset Places~\citep{zhou2017places}. With 62.5k images from 365 categories, its class cardinality ranges from 5 to 4,980.


\begin{table*}[t]
\centering
\renewcommand{\arraystretch}{1.1}
\resizebox{\linewidth}{!}{ 
  \begin{tabular}{llllllllllllll}
  \toprule
  \multicolumn{2}{c}{\multirow{2.5}{*}{\textbf{Method}}}  & \multicolumn{4}{c}{\textbf{ImageNet-LT}} & \multicolumn{4}{c}{\textbf{iNaturalist 18}} & \multicolumn{4}{c}{\textbf{Place-LT}} \\
  \cmidrule(r){3-6} \cmidrule(r){7-10} \cmidrule(r){11-14}
  \multicolumn{2}{c}{}     &\bf Overall &\bf Many &\bf Med. &\bf Few &\bf Overall &\bf Many &\bf Med. &\bf Few &\bf Overall &\bf Many &\bf Med. &\bf Few \\
  \midrule
  \multirow{4}{*}{\rotatebox[origin=c]{90}{Scratch}} & Baseline &60.9 & 72.9 & 56.8 & 41.4  & 76.1 & 78.5 & 76.9 & 74.6 & 39.9 & 43.0 & 40.5 & 33.3 \\
  ~& $+$ RD & 56.8 & 72.1 & 50.4 & 35.8 & 68.4 & 76.4 & 70.1 & 64.3 & 36.5 & 41.9 & 36.1 & 27.5 \\
  ~& $+$ SD &  64.9 & 73.4 & 62.1 & 50.6 & 76.8 & 78.6 & 77.2 & 75.7 & 41.6 & 43.4 & 42.1 & 36.9 \\
  ~ & $+$ {\it Ours} & 68.2$_\text{\bf\textcolor{darkgreen}{$\uparrow$7.3}}$ & 74.5$_\text{\bf\textcolor{darkgreen}{$\uparrow$1.6}}$ & 66.2$_\text{\bf\textcolor{darkgreen}{$\uparrow$9.4}}$ & 57.4$_\text{\bf\textcolor{darkgreen}{$\uparrow$16.0}}$ & 78.0$_\text{\bf\textcolor{darkgreen}{$\uparrow$1.9}}$ & 78.9$_\text{\bf\textcolor{darkgreen}{$\uparrow$0.4}}$ & 78.2$_\text{\bf\textcolor{darkgreen}{$\uparrow$1.3}}$ & 77.5$_\text{\bf\textcolor{darkgreen}{$\uparrow$2.9}}$ & 43.8$_\text{\bf\textcolor{darkgreen}{$\uparrow$3.9}}$ & 43.7$_\text{\bf\textcolor{darkgreen}{$\uparrow$0.7}}$ & 44.8$_\text{\bf\textcolor{darkgreen}{$\uparrow$4.3}}$ & 41.6$_\text{\bf\textcolor{darkgreen}{$\uparrow$8.3}}$ \\
  \midrule
  \multirow{4}{*}{\rotatebox[origin=c]{90}{CLIP}} & Baseline & 74.0 & 77.2 & 72.8 & 68.5 & 75.0 & 77.8 &76.5 &72.5 & 48.4 & 47.9 & 48.6 & 48.9 \\
  ~& $+$ RD & 68.8 & 75.4 & 67.4  & 55.2  & 67.7 & 75.1  & 69.8 & 63.1  & 43.3 & 45.1 &  43.4 & 40.2 \\
  ~& $+$ SD & 75.2 & 77.8 & 74.2 & 71.3 & 76.7 & 78.5 & 77.9 & 74.6 & 49.2 & 48.7 & 49.5 & 49.4 \\
  ~ & $+$ {\it Ours} & 77.3$_\text{\bf\textcolor{darkgreen}{$\uparrow$3.5}}$ & 79.1$_\text{\bf\textcolor{darkgreen}{$\uparrow$1.9}}$ & 76.8$_\text{\bf\textcolor{darkgreen}{$\uparrow$4.0}}$ & 74.1$_\text{\bf\textcolor{darkgreen}{$\uparrow$5.6}}$ & 78.5$_\text{\bf\textcolor{darkgreen}{$\uparrow$3.5}}$ & 79.5$_\text{\bf\textcolor{darkgreen}{$\uparrow$1.5}}$ & 79.3$_\text{\bf\textcolor{darkgreen}{$\uparrow$2.8}}$ & 77.3$_\text{\bf\textcolor{darkgreen}{$\uparrow$4.8}}$ & 50.5$_\text{\bf\textcolor{darkgreen}{$\uparrow$2.1}}$ & 50.0$_\text{\bf\textcolor{darkgreen}{$\uparrow$2.1}}$ & 51.0$_\text{\bf\textcolor{darkgreen}{$\uparrow$2.4}}$ & 50.2$_\text{\bf\textcolor{darkgreen}{$\uparrow$1.3}}$ \\
  \midrule
  \multirow{4}{*}{\rotatebox[origin=c]{90}{DINOv2}} & Baseline & 79.6 & 84.3 & 78.3 & 71.1 & 85.0 & 85.7 & 86.2 & 84.2 & 49.5 & 49.2 & 51.3 & 46.1 \\
  ~& $+$ RD & 77.2 & 83.3 & 75.7 & 65.4 & 75.4 & 82.3 & 76.1 & 72.6  & 45.2 & 47.2 & 45.2 &  41.3\\
  ~& $+$ SD & 80.5 & 83.8 & 79.8 & 73.4  & 85.9 &  85.8 & 86.5 & 85.0 & 49.9 & 49.3 & 51.6 & 47.3 \\
  ~ & $+$ {\it Ours} & 82.0$_\text{\bf\textcolor{darkgreen}{$\uparrow$2.4}}$ & 84.7$_\text{\bf\textcolor{darkgreen}{$\uparrow$0.4}}$ & 81.5$_\text{\bf\textcolor{darkgreen}{$\uparrow$3.2}}$ & 76.2$_\text{\bf\textcolor{darkgreen}{$\uparrow$5.1}}$ & 87.0$_\text{\bf\textcolor{darkgreen}{$\uparrow$2.0}}$ & 86.4$_\text{\bf\textcolor{darkgreen}{$\uparrow$0.7}}$ & 87.4$_\text{\bf\textcolor{darkgreen}{$\uparrow$1.2}}$ & 86.7$_\text{\bf\textcolor{darkgreen}{$\uparrow$2.5}}$ & 50.8$_\text{\bf\textcolor{darkgreen}{$\uparrow$1.3}}$ & 49.4$_\text{\bf\textcolor{darkgreen}{$\uparrow$0.2}}$ & 52.4$_\text{\bf\textcolor{darkgreen}{$\uparrow$1.1}}$ & 49.2$_\text{\bf\textcolor{darkgreen}{$\uparrow$3.1}}$ \\
  \bottomrule
\end{tabular}}
\caption{{\bf Quantitative results of the proposed method on three standard benchmarks.} For each dataset, we conduct three pre-training paradigms (training from scratch, CLIP, and DINOv2) to compare our method with baseline methods on accuracy ($\%$). In addition, we report the \textcolor{darkgreen}{relative improvement} of our method compared to the baseline method in each setting. RD denotes the random auxiliary data and SD is the data from our selected neighbor categories.}\label{tab:main_result} 
\vspace{-0.5cm}
\end{table*}

\subsection{Implementation Details} \label{subsec:implement_details}
We adopt ViT-Base~\citep{dosovitskiy2021vit} as the backbone. 
Our models are trained with the AdamW optimizer~\citep{AdamW} with $\beta_s= \{0.9, 0.95\}$, with an effective batch size of 512.
We train all models with ${\rm RandAug}(9, 0.5)$~\citep{Randaugment}, ${\rm Mixup}(0.8)$~\citep{ref:mixup_iclr2018} and ${\rm Cutmix}(1.0)$~\citep{ref:cutmix_iccv2019}. 
We set the maximum sampling number for each auxiliary category to 50 in each training epoch. 
For the ratio of neighbor category for head, medium, and tail class, we set to $1:\left[\frac{N_h}{N_m}\right]:\left[\frac{N_h}{N_t}\right]$, where $N_h$, $N_m$, and $N_t$ denote the total number of samples of head, medium, and tail classes, respectively. $\left[\cdot\right]$ stands for ceiling, which rounds a number up to the nearest integer.
Following LiVT~\citep{xu2023learning}, the training epochs for ImageNet-LT, iNaturalist, and Place-LT is set to 100, 100, and 30, respectively. 
The hyper-parameter $\lambda_s$ is set to 0.1. 
$\gamma_1$ and $\gamma_2$ are set to 0.7 and 0.98. 
See detailed implementation settings in the Appendix.

\begin{table}[t]
\vspace{0.2cm}
\centering
\centering
\setlength{\tabcolsep}{0.85ex} 
\renewcommand{\arraystretch}{1}
\small
\begin{tabular}{lccccc}
\toprule
\bf Methods &\bf Backbone  &\bf Overall &\bf Many &\bf Med. &\bf Few \\
\midrule
\multicolumn{6}{l}{\bf Training from scratch} \\
\midrule
LiVT~\citep{xu2023learning} & ViT-B  & 60.9 & 73.6 & 56.4 & 41.0 \\
LiVT$^\dagger$~\citep{xu2023learning} & ViT-B   & 65.2 & 73.7 &  62.8 & 49.8  \\
\emph{Ours}  & ViT-B  & \textbf{68.2} & \textbf{74.5} & \textbf{66.2} & \textbf{57.4} \\
\midrule
\multicolumn{6}{l}{\bf Fine-tuning pre-trained model (CLIP)} \\
\midrule
LIFT~\citep{shi2024longtail}  & ViT-B  & 77.0 & 80.2 & 76.1 & 71.5 \\
LIFT$^\dagger$~\citep{shi2024longtail}  & ViT-B  & 77.8 & 80.2 & 77.2 & 73.1 \\
\emph{Ours} & ViT-B  & \bf 78.8 & \bf 80.3 & \textbf{78.4} & \textbf{75.8} \\
\midrule
\multicolumn{6}{l}{\bf Fine-tuning pre-trained model (DINOv2)} \\
\midrule
Bal-CE~\citep{cui2019class} & ViT-B  & 79.6 & 84.3 & 78.3 & 71.1 \\
Bal-CE$^\dagger$~\citep{cui2019class} & ViT-B   & 80.5 & 83.8 & 79.8 & 73.4  \\
\emph{Ours} & ViT-B  & \textbf{82.0} & \textbf{84.7} & \textbf{81.5} & \textbf{76.2} \\
\bottomrule
\end{tabular}
\vspace{0.1cm}
\caption{{\bf Performance on ImageNet-LT.} We report accuracy ($\%$) of all methods under three pre-training paradigms. We also report the performance of adding the auxiliary data but without our method, which denotes by $^\dagger$.} 
\label{table:comp_imagenetlt}
\vspace{-0.6cm}
\end{table}

\subsection{Main Results} \label{subsec:results}

\noindent\textbf{Comparison with Baseline with Different Pre-training.}
We experiment with three different pre-training paradigms (\ie, random initialization, CLIP~\citep{radford2021clip}, and DINOv2~\citep{oquab2023dinov2}). 
The baseline applies Bal-CE~\citep{cui2019class} loss.
As shown in \cref{tab:main_result}, our method significantly improves the performance over the baseline on all three datasets, especially on fewer-shot classes. This improvement is also consistent and generalizes to a variety of pre-training strategies.
In particular, when the model is trained from scratch, we observe a significant performance boost on ImageNet-LT, with a 16.0\% increase in accuracy on the tail classes. 
A plausible explanation is that randomly initialized networks are more prone to overfitting on tail classes compared to large-scale pre-trained models. 
Our method effectively addresses this issue by utilizing neighbor categories.
Besides, even with pre-trained models as initialization, our approach consistently demonstrates satisfactory improvements. 
For example, when using DINOv2 as the backbone, we achieve performance improvements of 5.0\%, 2.5\%, and 3.1\% on the tail classes of ImageNet-LT, iNaturalist, and PlaceLT datasets, respectively, without compromising performance on the head classes.
This verifies our method's generalizability and effectiveness on long-tail datasets. 

\noindent\textbf{Fair comparison.} We also add auxiliary data to the baseline method. 
As shown in \cref{tab:main_result}, RD denotes the random auxiliary data and SD is the data from our selected neighbor categories. 
The results show that, the randomly auxiliary data significantly degrade the performance during the finetuning stage, and the selected neighbor categories can enhance performance. 
Moreover, when using our proposed methods with the neighbor categories, the peformance can be further boosted. 
These results validate the effectiveness of both the auxiliary data and our approach.

\noindent\textbf{Can Learning by Category Extrapolation Enhance the State-of-the-Art Methods?}
We conduct comprehensive experiments with existing SoTAs in \cref{table:comp_imagenetlt}, \cref{table:comp_inat18}, and \cref{table:placeslt}. 
Current methods can be generally categorized into two settings, \ie, training from scratch or adopting CLIP pre-training. 
We also present results obtained by DINOv2, in which we provide the results of Bal-CE~\citep{cui2019class} initialized by pre-trained weights from DINOv2. 
In each pre-training paradigm, we select a SOTA method, and add the same amount of auxiliary data on it, which is denoted by $^\dagger$.
Then we implement our proposed method based on the corresponding SoTA methods. 
%
%
The results show that 1) After adding the auxiliary data from neighbor categories, the performance increase. 2) When using the neighbor categories with our proposed methods, we can further enhance the performance. The potential reason is that our method focuses more effectively on learning the features of target classes, which avoids being overwhelmed by auxiliary categories.
Due to the space limitation, we show more results of previous methods trained with the auxiliary data in Appendix.

\begin{table}[t]
\centering
\setlength{\tabcolsep}{0.85ex} 
\renewcommand{\arraystretch}{1}
\small
\begin{tabular}{lccccc}
\toprule
\bf Method &\bf Backbone  &\bf Overall &\bf Many &\bf Med. &\bf Few \\
\midrule
\multicolumn{6}{l}{\bf Training from scratch} \\
\midrule
LiVT~\citep{xu2023learning} & ViT-B  & 76.1 & 78.9 & 76.5 & 74.8 \\
LiVT$^\dagger$~\citep{xu2023learning} & ViT-B   & 77.0 & 78.8 & 77.4 &  75.9 \\
\emph{Ours}  & ViT-B  & \textbf{78.0} & \textbf{78.9} & \textbf{78.2} & \textbf{77.5} \\
\midrule
\multicolumn{6}{l}{\bf Fine-tuning pre-trained model (CLIP)} \\
\midrule
LIFT~\citep{shi2024longtail}  & ViT-B  & 79.1 & 72.4 & 79.0 & 81.1 \\
LIFT$^\dagger$~\citep{shi2024longtail}  & ViT-B  & 79.5 & 72.9 & 79.4 & 81.3 \\
\emph{Ours} & ViT-B  & \textbf{80.9} & \textbf{79.6} & \textbf{80.1} & \textbf{82.1} \\
\midrule
\multicolumn{6}{l}{\bf Fine-tuning pre-trained model (DINOv2)} \\
\midrule
Bal-CE~\citep{cui2019class} & ViT-B  & 85.0 & 85.7 & 86.2 & 84.2 \\
Bal-CE$^\dagger$~\citep{cui2019class} & ViT-B   & 85.9 &  85.8 & 86.5 & 85.0  \\
\emph{Ours} & ViT-B & \textbf{87.0} & \textbf{86.4} & \textbf{87.4} & \textbf{86.7} \\
\bottomrule
\end{tabular}
\vspace{0.1cm}
\caption{{\bf Performance on iNaturalist 2018.} We report accuracy ($\%$) of all methods under three pre-training paradigms.} 
\label{table:comp_inat18}
\vspace{-0.6cm}
\end{table}

\begin{table}[t]
\centering
\setlength{\tabcolsep}{0.75ex} 
\renewcommand{\arraystretch}{1}
\small
\begin{tabular}{lccccc}
\toprule
\bf Method &\bf Backbone  &\bf Overall &\bf Many &\bf Med. &\bf Few \\
\midrule
\multicolumn{6}{l}{\bf Training from scratch } \\
\midrule
LiVT~\citep{xu2023learning} & ViT-B  & 40.8 & \textbf{48.1} & 40.6 & 27.5 \\
LiVT$^\dagger$~\citep{xu2023learning} & ViT-B  & 42.8 & 48.0 & 42.0 & 35.1 \\
\emph{Ours} & ViT-B  & \textbf{43.8} & 43.7 & \textbf{44.8} & \textbf{41.6} \\
\midrule
\multicolumn{6}{l}{\bf Fine-tuning pre-trained model (CLIP)} \\
\midrule
LIFT~\citep{shi2024longtail}  & ViT-B  & 51.5 & 51.3 & 52.2 & 50.5 \\
LIFT$^\dagger$~\citep{shi2024longtail} & ViT-B  & 51.8 & 51.5  & 52.4 & 51.1 \\
\emph{Ours} & ViT-B   & \bf 52.4 &  \bf 51.6 & \bf 53.0  & \bf 52.3 \\
\midrule
\multicolumn{6}{l}{\bf Fine-tuning pre-trained model (DINOv2)} \\
\midrule
Bal-CE~\citep{cui2019class} & ViT-B & 49.5 & 49.2 & 51.3 & 46.1 \\
Bal-CE$^\dagger$~\citep{cui2019class} & ViT-B  & 49.9 & 49.3 & 51.6 & 47.3\\
\emph{Ours} & ViT-B  & \bf 50.8 & \bf 49.4 & \bf 52.4  & \bf 49.2 \\
\bottomrule
\end{tabular}
\vspace{0.1cm}
\caption{{\bf Performance on Places-LT.} We report accuracy ($\%$) of all methods under three pre-training paradigms.} 
\label{table:placeslt}
\vspace{-0.5cm}
\end{table}

\begin{figure*}[t]
\centering
    \begin{subfigure}[t]{0.3333\textwidth}
        \centering
        \includegraphics[width=\textwidth]{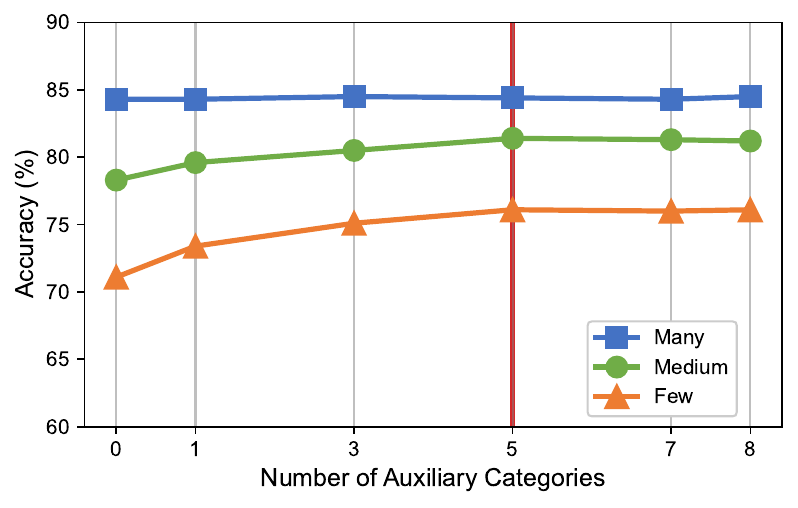}
        \caption{Effect of \#aux. categories.}
        \label{subfig:ablation_cls_num}
    \end{subfigure}\hfill%
    \begin{subfigure}[t]{0.3333\textwidth}
        \centering
        \includegraphics[width=\textwidth]{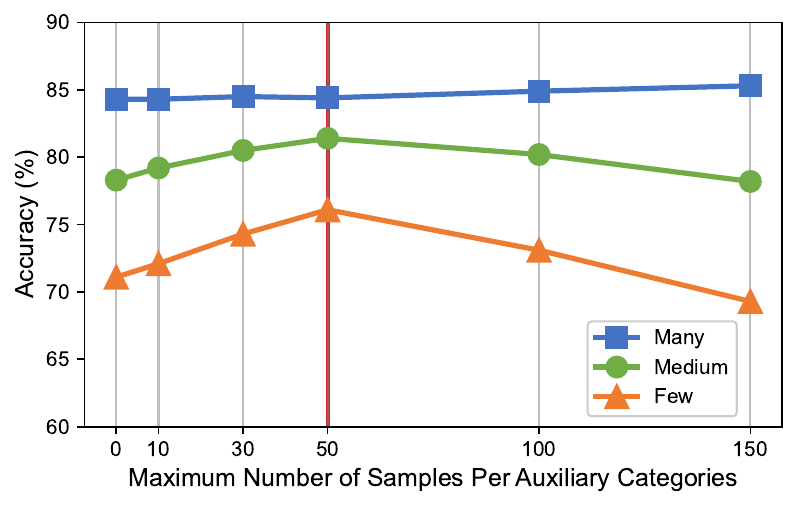}
        \caption{Effect of \#sample/aux. class.}
        \label{subfig:ablation_sample_num}
    \end{subfigure}\hfill%
    \begin{subfigure}[t]{0.3333\textwidth}
        \centering
        \includegraphics[width=\textwidth]{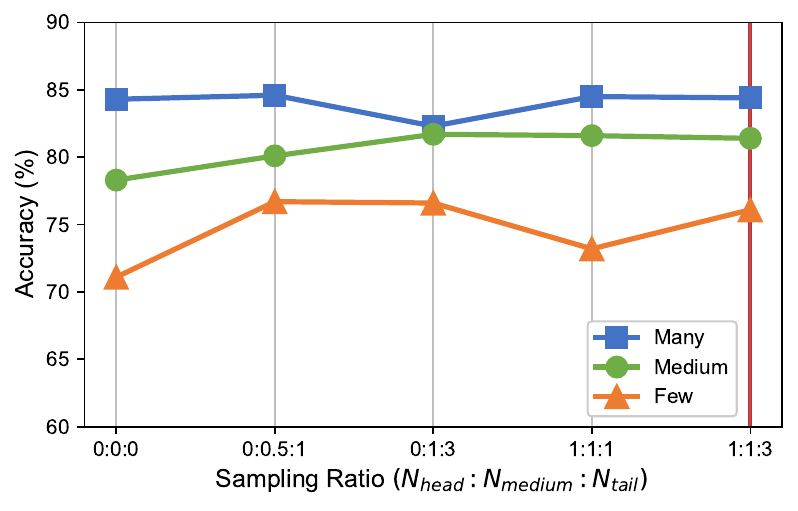}
        \caption{Effect of sampling ratio.}
        \label{subfig:ablation_sample_ratio}
    \end{subfigure}
\vspace{-0.1cm}
   \caption{\textbf{Ablation study on factors related to the curation of auxiliary dataset.} Experiments are conducted on ImageNet-LT~\citep{liu2019oltr}. Default options are marked in \textcolor{BrickRed}{red}.} \label{fig:ablation}
\vspace{-0.35cm}
\end{figure*}

\subsection{Ablation and Analysis} \label{subsec:ablation}

\noindent\textbf{Contributions of Individual Components.}
As shown in Tab. \red{6}, we evaluate the contribution of each component of the full method.
The baseline is BalCE with  DINOv2 pretraining. We conduct ablation experiments on ImageNet-LT. 
%
%
We replace the re-balancing loss (\cref{eq:bal-ce}) with the neighbor-silencing loss (\cref{eq:sil-ce}), obtaining improvements of 1.0\% and 1.9\% in the medium and tail categories, respectively. 
If we use the direct classifier instead of retraining the classfier by linear probing, the performance in the medium and tail categories increases to 79.2\% and 73.2\%, respectively.
The best performance is achieved when we do not re-train the classifier and instead directly utilize the classifier weights corresponding to the target categories.

The curation of the auxiliary dataset primarily involves three hyper-parameters: the number of auxiliary categories associated with a target category, the maximum number of samples per auxiliary class, and the proportion of the number of auxiliary categories for head ($\text{aux}_\text{head}$), medium ($\text{aux}_\text{medium}$), and tail classes ($\text{aux}_\text{tail}$), i.e. $\text{aux}_\text{head}:\text{aux}_\text{medium}:\text{aux}_\text{tail}$ (denoted as auxiliary sampling ratio for simplicity).
We will analyze these three hyper-parameters separately and fix the other two hyper-parameters individually. The default values for these three hyper-parameters are 5, 50, and 1:1:3, respectively.

\noindent\textbf{Number of Sampled Categories.}
\cref{subfig:ablation_cls_num} studies the effect of the number of auxiliary categories for each target class.
The optional values are set to $\{1, 3, 5, 7, 8\}$.
We can observe that as the number of neighbor categories increases, the performance gradually improves and finally saturates when approaching 5.

\noindent\textbf{Maximum Number of Sampled Instances Per Class.}
As shown in \cref{subfig:ablation_sample_num}, we study the effect of the number of samples per neighbor category.
The optional values are $\{10, 30, 50, 100, 150\}$.
If the number of samples collected for a class exceeds the limit, we randomly subsample it to the corresponding number; and if less, we keep them unchanged. 
It can be seen that as the limit increases to 50, the performance improves. 
However, when too many instances are included, the performance drops.
This can be attributed to an excessive number of samples from auxiliary classes, resulting in an overwhelming of these categories.

\begin{table}[t]
\centering
\setlength{\tabcolsep}{1.35ex} 
\renewcommand{\arraystretch}{1}
\small
\begin{tabular}{l|cccc}
\toprule
\bf Methods & \bf Many  & \bf Medium & \bf Few & \bf Overall  \\
\midrule
Baseline & 84.3 & 78.3 & 71.1 & 79.6 \\
+ Random Category & 83.3 & 75.7 & 65.4 & 77.2 \\
+ Neighbor Category & 83.8 & 79.8 & 73.4 & 80.5 \\
+ Neighbor Silencing & 84.3 & 80.8 & 75.3 & 81.4 \\
+ Direct Classifier & \bf 84.7 & \bf 81.5 & \bf 76.2 & \bf 82.0 \\
\bottomrule
\end{tabular}
\vspace{0.3ex}
\caption{{\bf Contributions of individual components.} Results are obtained on ImageNet-LT.} 
\label{tab:ablation}
\vspace{-0.6cm}
\end{table}

\noindent\textbf{Auxiliary Sampling Ratio.}
\cref{subfig:ablation_sample_ratio} studies the proportion of the number of auxiliary categories for head, medium, and tail classes.
When the ratio is 0:1:3, which indicates that the neighbor categories for many classes are removed, we can observe a performance degradation in many classes from 84.4\% to 82.3\%. 
This could be because, with only the addition of auxiliary data in the medium and few-shot categories, feature learning tends to skew towards these medium and few-shot categories.
Moreover, when we decrease the ratio on medium (ratio=1:0.5:3) and tail (ratio=1:1:1) classes, the performance degrades, respectively.

\begin{figure}[t]
\centering
\hspace{-0.2cm}
\includegraphics[width=\linewidth]{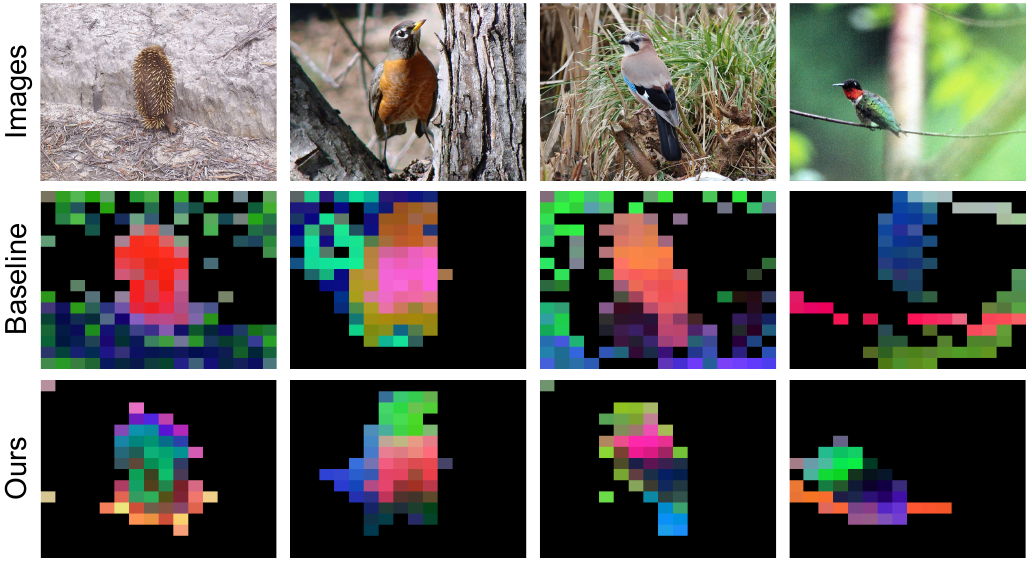}
\caption{{\bf PCA visualization of ``Tail'' images in ImageNet-LT.} Top-3 PCA components of features are mapped to RGB channels, and background is removed by thresholding the first component.} \label{fig:fea-maps}
\vspace{-0.2cm}
\end{figure}

\noindent\textbf{Visualization.}
\cref{fig:fea-maps} shows the top-3 PCA components of images sampled from ``Tail'' classes of ImageNet-LT, where each component is mapped to an RGB channel, and the background is removed by thresholding the first PCA component.
Both the baseline~\citep{cui2019class} and our method adopt DINOv2 pre-training.
While the baseline finds it hard to locate the object of interest, our method clearly captures better objectness despite the scarcity of ``Tail'' images.


\section{Related Works}

\noindent \textbf{Re-Balancing Long-Tail Learning.} \label{subsec:rebalance}
Class-level re-balancing methods include oversampling training samples from tail classes~\citep{chawla2002smote}, under-sampling data points from head classes~\citep{liu2006exploratory}, and re-weighting the loss values or gradients based on label frequencies~\citep{cao2019ldam,cui2019class} or model's predictions~\citep{lin2017focal}.
Classifier re-balancing mechanisms are based on the finding that uniform sampling on the whole dataset during training benefits representation learning but leads to a biased classifier, so they design specific algorithms to adjust the classifier during or after the representation learning phase~\citep{zhou2020bbn,kang2020decoupling}.


\noindent \textbf{Data Augmentation for Long-Tail Learning.} \label{subsec:data_aug}
Spatial augmentation methods have performed satisfactorily for representation learning.
Among these approaches, Cutout~\citep{ref:cutout_2017} removes random regions, CutMix~\citep{ref:cutmix_iccv2019} fills the removed regions with patches from other images, and Mixup series~\citep{ref:mixup_iclr2018,ref:manifoldmixup_2019,summers2019improvedmixup} performs convex combination between images.
Since data augmentation is closely related to oversampling, it is also adopted by recent long-tail recognition literature~\citep{zhou2020bbn,zhong2021mislas}.
These techniques, however, are adopted directly while overlooking special data distributions in long-tail learning.
Recently, Remix~\citep{chou2020remix} was proposed in favor of the minority classes when mixing samples. Yet, this is still bounded by existing classes. 
Unlike above, our method samples images from open-set distributions and could greatly benefit from higher data diversity.


\noindent \textbf{Auxiliary Resources for Long-Tail Learning.} \label{subsec:aux_resource}
Previous efforts mainly lie in refining representations with fixed external image features encoded by pre-trained models~\citep{long2022rac,iscen2023improving}.
The external data could be either the training dataset~\citep{long2022rac} or crawled from the web~\citep{iscen2023improving}, and the fusing process could be either non-parametric~\citep{long2022rac} or learned in an attentive fashion~\citep{iscen2023improving}.
Besides images, another line~\citep{tian2022vl} is to leverage external textual descriptors encoded by vision-language models~\citep{radford2021clip}.
Our method, instead, poses a clear contrast by explicitly introducing external open-set data into a clean training pipeline and is not dependent on any foundation model.
There is also a recent work in self-supervised learning that shares the idea of crawling visually-similar data for task-specific improvements~\citep{li2023internet}. Instead, our work places a special focus on long-tail learning.

\section{Concluding Remarks} \label{sec:conclu}
This paper introduces category extrapolation, which leverages diverse open-set images crawled from the web to enhance closed-set long-tail learning. 
In addition to a clean and decent method that shows superior performance on ``Medium'' and ``Few'' splits across standard benchmarks, we also provide instrumental guidance on when the auxiliary data helps most and empirical explanations on how they help shape the feature manifold through visualizations.
We hope our research will attract more researchers to consider how to leverage additional data to address the pervasive problem in long-tail learning.
Related research topics could include (i) what kind of additional data is more compatible with target datasets and (ii) how to take the additional data in conjunction with target datasets for training.

\clearpage
{
    \small
    \bibliographystyle{ieeenat_fullname}
    \bibliography{main}

\begin{thebibliography}{47}
\providecommand{\natexlab}[1]{#1}
\providecommand{\url}[1]{\texttt{#1}}
\expandafter\ifx\csname urlstyle\endcsname\relax
  \providecommand{\doi}[1]{doi: #1}\else
  \providecommand{\doi}{doi: \begingroup \urlstyle{rm}\Url}\fi

\bibitem[Alshammari et~al.(2022)Alshammari, Wang, Ramanan, and Kong]{LTR-WD}
Shaden Alshammari, Yu-Xiong Wang, Deva Ramanan, and Shu Kong.
\newblock Long-tailed recognition via weight balancing.
\newblock In \emph{CVPR}, 2022.

\bibitem[Cao et~al.(2019)Cao, Wei, Gaidon, Arechiga, and Ma]{cao2019ldam}
Kaidi Cao, Colin Wei, Adrien Gaidon, Nikos Arechiga, and Tengyu Ma.
\newblock Learning imbalanced datasets with label-distribution-aware margin loss.
\newblock In \emph{NeurIPS}, 2019.

\bibitem[Chawla et~al.(2002)Chawla, Bowyer, Hall, and Kegelmeyer]{chawla2002smote}
Nitesh~V Chawla, Kevin~W Bowyer, Lawrence~O Hall, and W~Philip Kegelmeyer.
\newblock Smote: synthetic minority over-sampling technique.
\newblock \emph{JAIR}, 2002.

\bibitem[Chou et~al.(2020)Chou, Chang, Pan, Wei, and Juan]{chou2020remix}
Hsin-Ping Chou, Shih-Chieh Chang, Jia-Yu Pan, Wei Wei, and Da-Cheng Juan.
\newblock Remix: rebalanced mixup.
\newblock In \emph{ECCV Workshops}, 2020.

\bibitem[Cubuk et~al.(2020)Cubuk, Zoph, Shlens, and Le]{Randaugment}
Ekin~D Cubuk, Barret Zoph, Jonathon Shlens, and Quoc~V Le.
\newblock Randaugment: Practical automated data augmentation with a reduced search space.
\newblock In \emph{CVPR Workshops}, 2020.

\bibitem[Cui et~al.(2021)Cui, Zhong, Liu, Yu, and Jia]{cui2021paco}
Jiequan Cui, Zhisheng Zhong, Shu Liu, Bei Yu, and Jiaya Jia.
\newblock Parametric contrastive learning.
\newblock In \emph{ICCV}, 2021.

\bibitem[Cui et~al.(2019)Cui, Jia, Lin, Song, and Belongie]{cui2019class}
Yin Cui, Menglin Jia, Tsung-Yi Lin, Yang Song, and Serge Belongie.
\newblock Class-balanced loss based on effective number of samples.
\newblock In \emph{CVPR}, 2019.

\bibitem[DeVries and Taylor(2017)]{ref:cutout_2017}
Terrance DeVries and Graham~W Taylor.
\newblock Improved regularization of convolutional neural networks with cutout.
\newblock \emph{arXiv:1708.04552}, 2017.

\bibitem[Dong et~al.(2023)Dong, Zhou, Yan, and Zuo]{dong2023lpt}
Bowen Dong, Pan Zhou, Shuicheng Yan, and Wangmeng Zuo.
\newblock {LPT}: Long-tailed prompt tuning for image classification.
\newblock In \emph{ICLR}, 2023.

\bibitem[Dosovitskiy et~al.(2021)Dosovitskiy, Beyer, Kolesnikov, Weissenborn, Zhai, Unterthiner, Dehghani, Minderer, Heigold, Gelly, Uszkoreit, and Houlsby]{dosovitskiy2021vit}
Alexey Dosovitskiy, Lucas Beyer, Alexander Kolesnikov, Dirk Weissenborn, Xiaohua Zhai, Thomas Unterthiner, Mostafa Dehghani, Matthias Minderer, Georg Heigold, Sylvain Gelly, Jakob Uszkoreit, and Neil Houlsby.
\newblock An image is worth 16x16 words: Transformers for image recognition at scale.
\newblock In \emph{ICLR}, 2021.

\bibitem[He et~al.(2016)He, Zhang, Ren, and Sun]{he2016deep}
Kaiming He, Xiangyu Zhang, Shaoqing Ren, and Jian Sun.
\newblock Deep residual learning for image recognition.
\newblock In \emph{CVPR}, 2016.

\bibitem[He et~al.(2021)He, Wu, and Wei]{he2021dive}
Yin-Yin He, Jianxin Wu, and Xiu-Shen Wei.
\newblock Distilling virtual examples for long-tailed recognition.
\newblock In \emph{ICCV}, 2021.

\bibitem[Iscen et~al.(2023)Iscen, Fathi, and Schmid]{iscen2023improving}
Ahmet Iscen, Alireza Fathi, and Cordelia Schmid.
\newblock Improving image recognition by retrieving from web-scale image-text data.
\newblock In \emph{CVPR}, 2023.

\bibitem[Kang et~al.(2020)Kang, Xie, Rohrbach, Yan, Gordo, Feng, and Kalantidis]{kang2020decoupling}
Bingyi Kang, Saining Xie, Marcus Rohrbach, Zhicheng Yan, Albert Gordo, Jiashi Feng, and Yannis Kalantidis.
\newblock Decoupling representation and classifier for long-tailed recognition.
\newblock In \emph{ICLR}, 2020.

\bibitem[Krasin et~al.(2017)Krasin, Duerig, Alldrin, Ferrari, Abu-El-Haija, Kuznetsova, Rom, Uijlings, Popov, Veit, Belongie, Gomes, Gupta, Sun, Chechik, Cai, Feng, Narayanan, and Murphy]{openimages}
Ivan Krasin, Tom Duerig, Neil Alldrin, Vittorio Ferrari, Sami Abu-El-Haija, Alina Kuznetsova, Hassan Rom, Jasper Uijlings, Stefan Popov, Andreas Veit, Serge Belongie, Victor Gomes, Abhinav Gupta, Chen Sun, Gal Chechik, David Cai, Zheyun Feng, Dhyanesh Narayanan, and Kevin Murphy.
\newblock Openimages: A public dataset for large-scale multi-label and multi-class image classification.
\newblock \emph{Dataset available from https://github.com/openimages}, 2017.

\bibitem[Li et~al.(2023)Li, Brown, Efros, and Pathak]{li2023internet}
Alexander~C Li, Ellis Brown, Alexei~A Efros, and Deepak Pathak.
\newblock Internet explorer: Targeted representation learning on the open web.
\newblock In \emph{ICML}, 2023.

\bibitem[Lin et~al.(2017)Lin, Goyal, Girshick, He, and Doll{\'a}r]{lin2017focal}
Tsung-Yi Lin, Priya Goyal, Ross Girshick, Kaiming He, and Piotr Doll{\'a}r.
\newblock Focal loss for dense object detection.
\newblock In \emph{ICCV}, 2017.

\bibitem[Liu et~al.(2025)Liu, Wen, Zhao, Chen, and Qi]{liu2025can}
Jiahui Liu, Xin Wen, Shizhen Zhao, Yingxian Chen, and Xiaojuan Qi.
\newblock Can ood object detectors learn from foundation models?
\newblock In \emph{ECCV}, 2025.

\bibitem[Liu et~al.(2006)Liu, Wu, and Zhou]{liu2006exploratory}
Xu-Ying Liu, Jianxin Wu, and Zhi-Hua Zhou.
\newblock Exploratory undersampling for class-imbalance learning.
\newblock In \emph{ICDM}, 2006.

\bibitem[Liu et~al.(2019)Liu, Miao, Zhan, Wang, Gong, and Yu]{liu2019oltr}
Ziwei Liu, Zhongqi Miao, Xiaohang Zhan, Jiayun Wang, Boqing Gong, and Stella~X Yu.
\newblock Large-scale long-tailed recognition in an open world.
\newblock In \emph{CVPR}, 2019.

\bibitem[Long et~al.(2022)Long, Yin, Ajanthan, Nguyen, Purkait, Garg, Blair, Shen, and van~den Hengel]{long2022rac}
Alexander Long, Wei Yin, Thalaiyasingam Ajanthan, Vu Nguyen, Pulak Purkait, Ravi Garg, Alan Blair, Chunhua Shen, and Anton van~den Hengel.
\newblock Retrieval augmented classification for long-tail visual recognition.
\newblock In \emph{CVPR}, 2022.

\bibitem[Loshchilov and Hutter(2019)]{AdamW}
Ilya Loshchilov and Frank Hutter.
\newblock Decoupled weight decay regularization.
\newblock In \emph{ICLR}, 2019.

\bibitem[Maxime et~al.(2023)Maxime, Timothée, Théo, Huy, Marc, Vasil, Pierre, Daniel, Francisco, Alaaeldin, Mahmoud, Nicolas, Wojciech, Russell, Po-Yao, Shang-Wen, Ishan, Michael, Vasu, Gabriel, Hu, Hervé, Julien, Patrick, Armand, and Piotr]{oquab2023dinov2}
Oquab Maxime, Darcet Timothée, Moutakanni Théo, Vo Huy, Szafraniec Marc, Khalidov Vasil, Fernandez Pierre, Haziza Daniel, Massa Francisco, El-Nouby Alaaeldin, Assran Mahmoud, Ballas Nicolas, Galuba Wojciech, Howes Russell, Huang Po-Yao, Li Shang-Wen, Misra Ishan, Rabbat Michael, Sharma Vasu, Synnaeve Gabriel, Xu Hu, Jegou Hervé, Mairal Julien, Labatut Patrick, Joulin Armand, and Bojanowski Piotr.
\newblock Dinov2: Learning robust visual features without supervision.
\newblock \emph{arXiv:2304.07193}, 2023.

\bibitem[McInnes et~al.(2020)McInnes, Healy, and Melville]{mcinnes2020umap}
Leland McInnes, John Healy, and James Melville.
\newblock Umap: Uniform manifold approximation and projection for dimension reduction.
\newblock \emph{arXiv:1802.03426}, 2020.

\bibitem[OpenAI(2023)]{openai2023gpt4}
OpenAI.
\newblock Gpt-4 technical report.
\newblock \emph{arXiv:2303.08774}, 2023.

\bibitem[Parisot et~al.(2022)Parisot, Esperan{\c{c}}a, McDonagh, Madarasz, Yang, and Li]{CKT}
Sarah Parisot, Pedro~M Esperan{\c{c}}a, Steven McDonagh, Tamas~J Madarasz, Yongxin Yang, and Zhenguo Li.
\newblock Long-tail recognition via compositional knowledge transfer.
\newblock In \emph{CVPR}, 2022.

\bibitem[Park et~al.(2022)Park, Hong, Heo, Yun, and Choi]{CMO}
Seulki Park, Youngkyu Hong, Byeongho Heo, Sangdoo Yun, and Jin~Young Choi.
\newblock The majority can help the minority: Context-rich minority oversampling for long-tailed classification.
\newblock In \emph{CVPR}, 2022.

\bibitem[Radford et~al.(2021)Radford, Kim, Hallacy, Ramesh, Goh, Agarwal, Sastry, Askell, Mishkin, Clark, et~al.]{radford2021clip}
Alec Radford, Jong~Wook Kim, Chris Hallacy, Aditya Ramesh, Gabriel Goh, Sandhini Agarwal, Girish Sastry, Amanda Askell, Pamela Mishkin, Jack Clark, et~al.
\newblock Learning transferable visual models from natural language supervision.
\newblock In \emph{ICML}, 2021.

\bibitem[Ridnik et~al.(2021)Ridnik, Ben-Baruch, Noy, and Zelnik-Manor]{ridnik2021imagenet21k}
Tal Ridnik, Emanuel Ben-Baruch, Asaf Noy, and Lihi Zelnik-Manor.
\newblock Imagenet-21k pretraining for the masses, 2021.

\bibitem[Russakovsky et~al.(2015)Russakovsky, Deng, Su, Krause, Satheesh, Ma, Huang, Karpathy, Khosla, Bernstein, Berg, and Fei-Fei]{Imagenet}
Olga Russakovsky, Jia Deng, Hao Su, Jonathan Krause, Sanjeev Satheesh, Sean Ma, Zhiheng Huang, Andrej Karpathy, Aditya Khosla, Michael Bernstein, Alexander~C. Berg, and Li Fei-Fei.
\newblock {ImageNet Large Scale Visual Recognition Challenge}.
\newblock \emph{IJCV}, 2015.

\bibitem[Samuel and Chechik(2021)]{samuel2021dro}
Dvir Samuel and Gal Chechik.
\newblock Distributional robustness loss for long-tail learning.
\newblock In \emph{ICCV}, 2021.

\bibitem[Shi et~al.(2024)Shi, Wei, Zhou, Shao, Han, and Li]{shi2024longtail}
Jiang-Xin Shi, Tong Wei, Zhi Zhou, Jie-Jing Shao, Xin-Yan Han, and Yu-Feng Li.
\newblock Long-tail learning with foundation model: Heavy fine-tuning hurts.
\newblock In \emph{ICML}, 2024.

\bibitem[Simonyan and Zisserman(2015)]{karen2015vggnet}
Karen Simonyan and Andrew Zisserman.
\newblock Very deep convolutional networks for large-scale image recognition.
\newblock In \emph{ICLR}, 2015.

\bibitem[Summers and Dinneen(2019)]{summers2019improvedmixup}
Cecilia Summers and Michael~J Dinneen.
\newblock Improved mixed-example data augmentation.
\newblock In \emph{WACV}, 2019.

\bibitem[Tian et~al.(2022)Tian, Wang, Zhu, Dai, and Qiao]{tian2022vl}
Changyao Tian, Wenhai Wang, Xizhou Zhu, Jifeng Dai, and Yu Qiao.
\newblock {VL-LTR:} learning class-wise visual-linguistic representation for long-tailed visual recognition.
\newblock In \emph{ECCV}, 2022.

\bibitem[Van~Horn et~al.(2018)Van~Horn, Mac~Aodha, Song, Cui, Sun, Shepard, Adam, Perona, and Belongie]{van2018inat}
Grant Van~Horn, Oisin Mac~Aodha, Yang Song, Yin Cui, Chen Sun, Alex Shepard, Hartwig Adam, Pietro Perona, and Serge Belongie.
\newblock The inaturalist species classification and detection dataset.
\newblock In \emph{CVPR}, 2018.

\bibitem[Verma et~al.(2019)Verma, Lamb, Beckham, Najafi, Mitliagkas, Lopez-Paz, and Bengio]{ref:manifoldmixup_2019}
Vikas Verma, Alex Lamb, Christopher Beckham, Amir Najafi, Ioannis Mitliagkas, David Lopez-Paz, and Yoshua Bengio.
\newblock Manifold mixup: Better representations by interpolating hidden states.
\newblock In \emph{ICML}, 2019.

\bibitem[Wang et~al.(2021)Wang, Lian, Miao, Liu, and Yu]{wang2021ride}
Xudong Wang, Long Lian, Zhongqi Miao, Ziwei Liu, and Stella~X Yu.
\newblock Long-tailed recognition by routing diverse distribution-aware experts.
\newblock In \emph{ICLR}, 2021.

\bibitem[Xiang et~al.(2020)Xiang, Ding, Han, et~al.]{LFME}
Liuyu Xiang, Guiguang Ding, Jungong Han, et~al.
\newblock Learning from multiple experts: Self-paced knowledge distillation for long-tailed classification.
\newblock In \emph{ECCV}, 2020.

\bibitem[Xu et~al.(2023)Xu, Liu, Yang, Chai, and Yuan]{xu2023learning}
Zhengzhuo Xu, Ruikang Liu, Shuo Yang, Zenghao Chai, and Chun Yuan.
\newblock Learning imbalanced data with vision transformers.
\newblock In \emph{CVPR}, 2023.

\bibitem[Yu et~al.(2022)Yu, Guo, Zhang, Fan, Wang, and Cheng]{IDR}
Sihao Yu, Jiafeng Guo, Ruqing Zhang, Yixing Fan, Zizhen Wang, and Xueqi Cheng.
\newblock A re-balancing strategy for class-imbalanced classification based on instance difficulty.
\newblock In \emph{CVPR}, 2022.

\bibitem[Yun et~al.(2019)Yun, Han, Oh, Chun, Choe, and Yoo]{ref:cutmix_iccv2019}
Sangdoo Yun, Dongyoon Han, Seong~Joon Oh, Sanghyuk Chun, Junsuk Choe, and Youngjoon Yoo.
\newblock Cutmix: Regularization strategy to train strong classifiers with localizable features.
\newblock In \emph{ICCV}, 2019.

\bibitem[Zhang et~al.(2018)Zhang, Cisse, Dauphin, and Lopez-Paz]{ref:mixup_iclr2018}
Hongyi Zhang, Moustapha Cisse, Yann~N. Dauphin, and David Lopez-Paz.
\newblock mixup: Beyond empirical risk minimization.
\newblock In \emph{ICLR}, 2018.

\bibitem[Zhong et~al.(2021)Zhong, Cui, Liu, and Jia]{zhong2021mislas}
Zhisheng Zhong, Jiequan Cui, Shu Liu, and Jiaya Jia.
\newblock Improving calibration for long-tailed recognition.
\newblock In \emph{CVPR}, 2021.

\bibitem[Zhou et~al.(2017)Zhou, Lapedriza, Khosla, Oliva, and Torralba]{zhou2017places}
Bolei Zhou, Agata Lapedriza, Aditya Khosla, Aude Oliva, and Antonio Torralba.
\newblock Places: A 10 million image database for scene recognition.
\newblock \emph{IEEE TPAMI}, 2017.

\bibitem[Zhou et~al.(2020)Zhou, Cui, Wei, and Chen]{zhou2020bbn}
Boyan Zhou, Quan Cui, Xiu-Shen Wei, and Zhao-Min Chen.
\newblock {BBN}: Bilateral-branch network with cumulative learning for long-tailed visual recognition.
\newblock In \emph{CVPR}, 2020.

\bibitem[Zhu et~al.(2022)Zhu, Wang, Chen, Chen, and Jiang]{zhu2022balanced}
Jianggang Zhu, Zheng Wang, Jingjing Chen, Yi-Ping~Phoebe Chen, and Yu-Gang Jiang.
\newblock Balanced contrastive learning for long-tailed visual recognition.
\newblock In \emph{CVPR}, 2022.

\end{thebibliography}
}

\end{document}


\maketitle

\begin{abstract}
Balancing training on long-tail data distributions remains a long-standing challenge in deep learning. While methods such as re-weighting and re-sampling help alleviate the imbalance issue, limited sample diversity continues to hinder models from learning robust and generalizable feature representations, particularly for tail classes.
In contrast to existing methods, we offer a novel perspective on long-tail learning, inspired by an observation: datasets with finer granularity tend to be less affected by data imbalance. In this paper, we investigate this phenomenon through both quantitative and qualitative studies, showing that increased granularity enhances the generalization of learned features in tail categories. 
Motivated by these findings, we propose a method to increase dataset granularity through category extrapolation. Specifically, we introduce open-set fine-grained classes that are related to existing ones, aiming to enhance representation learning for both head and tail classes. 
To automate the curation of auxiliary data, we leverage large language models (LLMs) as knowledge bases to search for auxiliary categories and retrieve relevant images through web crawling. To prevent the overwhelming presence of auxiliary classes from disrupting training, we introduce a neighbor-silencing loss that encourages the model to focus on class discrimination within the target dataset. 
During inference, the classifier weights for auxiliary categories are masked out, leaving only the target class weights for use.  Extensive experiments on three standard long-tail benchmarks demonstrate the effectiveness of our approach, notably outperforming strong baseline methods that use the same amount of data. The code will be made publicly available. 
\end{abstract}

\section{Introduction}

Deep models have shown extraordinary performance on large-scale curated datasets~\citep{he2016deep,karen2015vggnet,dosovitskiy2021vit}.
But when dealing with real-world applications, they generally face highly imbalanced (\eg, long-tailed) data distribution: instances are dominated by a few head classes, and most classes only possess a few images~\citep{wang2021ride,he2021dive,LFME,dong2023lpt}.
Learning in such an imbalanced setting is challenging as the instance-rich (or head) classes dominate the training procedure~\citep{cui2021paco,samuel2021dro,LTR-WD,zhong2021mislas}.
Without considering this situation, models tend to classify tailed class samples as similar head categories, leading to significant performance degradation on tail categories~\citep{IDR,CMO,CKT,zhu2022balanced}. 

\begin{figure}[t]
\centering
\includegraphics[width=1\linewidth]{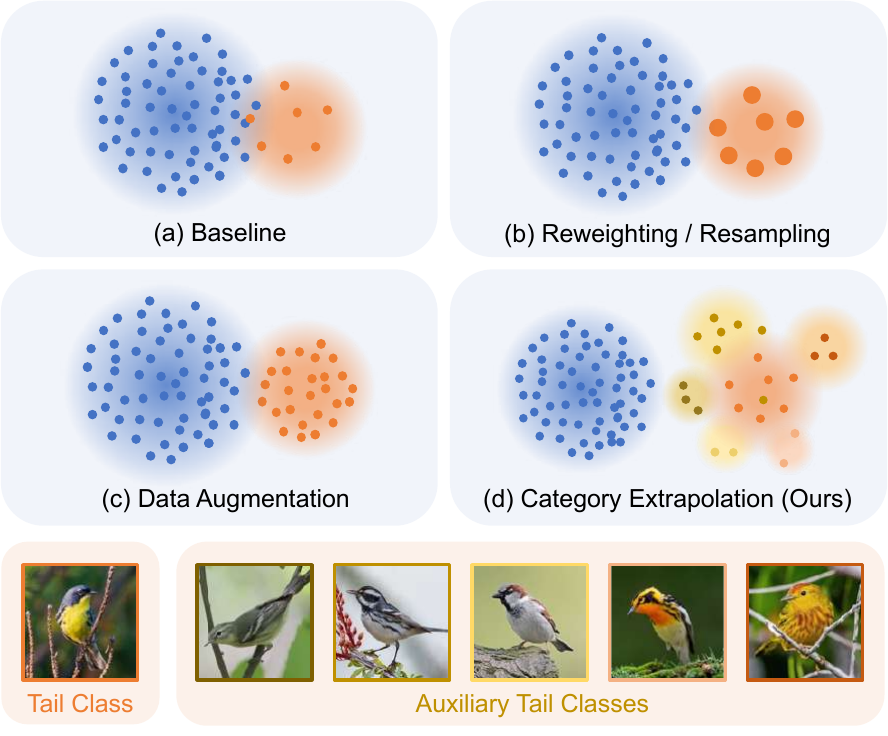}
\caption{{\bf Holistic comparison to previous philosophy.} (a) Data imbalance between \head{head} and \tail{tail} classes makes biased features; (b, c): Previous methods are still bounded by existing known classes; (d) We instead seek help from \aux{auxiliary} open-set data.} \label{fig:method-compare}
\end{figure}

Existing works tackle challenges in long-tail learning from various perspectives.
An earlier stream is to re-balance the learning signal (\eg, re-weighting~\citep{cui2019class} and re-sampling~\citep{chawla2002smote}). 
Yet, they inevitably face the scarcity of data and suffer from over-fitting on tail classes (\cref{fig:method-compare}\red{b}). 
Another straightforward fix is to augment training samples into diverse ones through image transformations~\citep{ref:cutout_2017,ref:mixup_iclr2018,ref:cutmix_iccv2019,chou2020remix}. 
These methods typically increase the loss weights or enhance the sample diversity of tail classes to balance representation learning (\cref{fig:method-compare}\red{c}). 
Despite advances, limited sample diversity still constrains the ability to generalize the learned features. Additionally, improvements in tail class performance are often accompanied by a decline in head class performance. 
This limitation motivates us to investigate what factors contribute to generalizable feature learning in long-tail settings. Our insight is inspired by a common, yet counterintuitive, phenomenon observed in existing benchmarks: despite being more imbalanced than ImageNet-LT~\citep{liu2019oltr}, iNat18~\citep{van2018inat} achieves nearly balanced performance (see \cref{tab:avg_perf}). 
This observation raises the question: \textit{Does granularity play a role in the performance balance of long-tail learning?}

\begin{table}[h]
\centering
\tablestylesmaller{3.3pt}{1}
\begin{tabular}{lllllccc}
\toprule
Dataset  & \#Class & \#Train & Granul.  & Imb. Factor $\beta$ &  Many  &  Med.  &  Few \\
\cmidrule(r){1-5} \cmidrule{6-8}
IN-LT    &  1000   &  116K   & Coarse  & 1280/5=256     & 68.2   & 56.8     & 41.6  \\
iNat18   &  8142   &  438K   & Fine    & 1000/2=500     & 70.3   & 71.3     & 70.2  \\
\bottomrule
\end{tabular}
\vspace{.0em}
\caption{Average performance of previous methods.} \label{tab:avg_perf}
\vspace{-1em}
\end{table}

\begin{figure*}[t]

\begin{center}
    \begin{subfigure}{0.24\textwidth}
        \includegraphics[width=\textwidth]{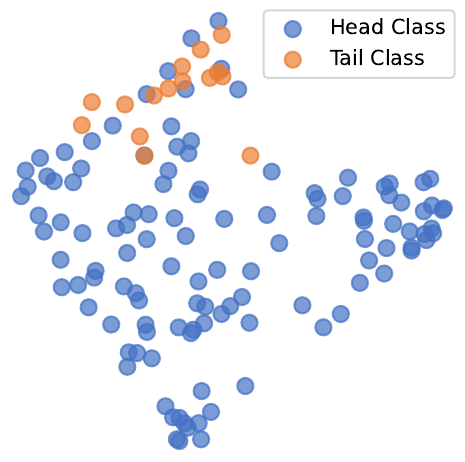}
        \caption{Raw feature space (train).}
        \label{fig:feature_vis_a}
    \end{subfigure}
    \hfill
    \begin{subfigure}{0.24\textwidth}
        \includegraphics[width=\textwidth]{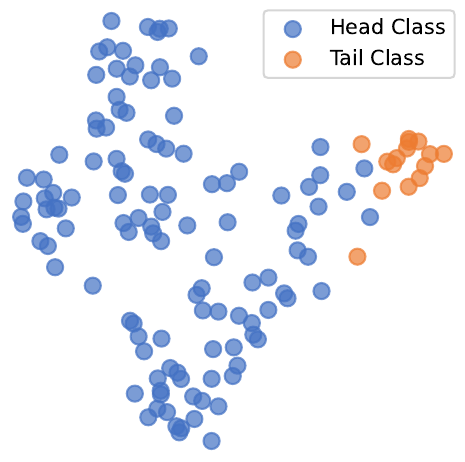}
        \caption{Baseline after training (train).}
        \label{fig:feature_vis_b}
    \end{subfigure}
    \hfill
    \begin{subfigure}{0.24\textwidth}
        \includegraphics[width=\textwidth]{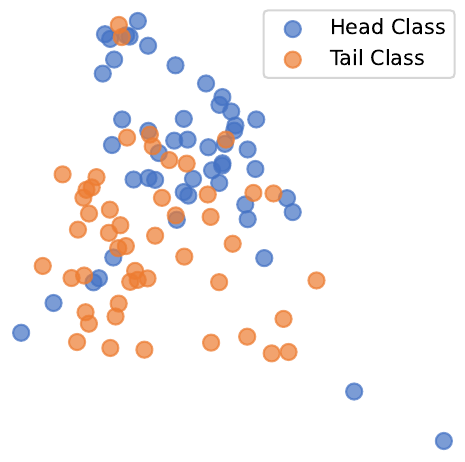}
        \caption{Baseline after training (val).}
        \label{fig:feature_vis_c}
    \end{subfigure}
    \hfill
    \begin{subfigure}{0.24\textwidth}
        \includegraphics[width=\textwidth]{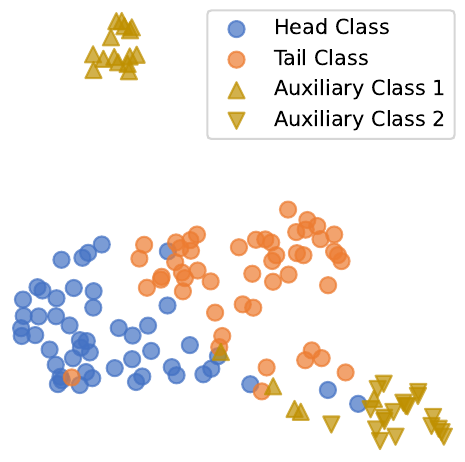}
        \caption{After training w/ aux. class (val).}
        \label{fig:feature_vis_d}
    \end{subfigure}
    \hfill
   \caption{\textbf{Feature visualization of confusing \head{head} and \tail{tail} classes by UMAP~\citep{mcinnes2020umap} on ImageNet-LT~\citep{liu2019oltr}.} (a) Raw feature space of training data by DINOv2~\citep{oquab2023dinov2}; (b) Feature space of training data after the training phase; (c) The baseline (re-weighting) shows poor generalization on validation data; (d) Adding auxiliary categories condenses clusters and improves separation.} \label{fig:feature_vis}
\end{center}
\vspace{-0.7cm}
\end{figure*}

To investigate this further, we conducted a pilot study (see \cref{subsec:granularity_effect}) using a larger data pool and controlled experiments to verify this phenomenon. We found that datasets with finer granularity are less affected by data imbalance.
Feature visualizations (see \cref{fig:feature_vis}) reveal that, despite a long-tail distribution, datasets with finer granularity enable the model to learn more generalized representations. This discovery motivates us to explore \textit{altering data distribution by introducing open-set categories to increase the granularity of data for long-tail learning.}


At the core of our approach is the idea of augmenting training data with fine-grained categories related to the original ones, thereby increasing granularity (\cref{fig:method-compare}\red{d}). 
To acquire auxiliary data, we establish a fully automated data crawling pipeline powered by the knowledge of large language models (LLMs). 
Specifically, for each class to be expanded, we query an LLM for $k$ fine-grained auxiliary classes, then retrieve corresponding images from the web based on these class names (\cref{fig:method-search}). 
The crawled data are subsequently integrated with the original dataset for model training.   
During training, we introduce a neighbor-silencing loss to enhance discrimination between confusing classes, prevent the model from being overwhelmed by auxiliary classes, and ensure alignment with the objectives of the testing phase. After training, the classifier by simply masking out the auxiliary classes demonstrates strong performance without the need for additional classifier re-balancing, as required in previous methods~\citep{kang2020decoupling,zhou2020bbn}.

Intuitively, our method could be interpreted as \textit{category extrapolation}. These augmented categories complete the learning signal, which may fill the gap between originally distinct classes, encourage continuity and smoothness of the feature manifold, and allow better generalization of representations across classes. In terms of classification, samples of auxiliary classes take up the neighborhood of existing classes, thus explicitly enlarging the margin between them and encouraging discriminability. Empirically, we indeed observe tighter clusters and better separation in-between (\cref{fig:feature_vis_d}).

Our major contributions are summarized as follows: 
\begin{itemize}
\item 
We explore the effect of granularity on the performance balance in long-tail learning, which motivate us to introduce neighbor classes to increase the granularity and facilitate representation learning for both head and tail classes. 
\item We propose a neighbor-silencing learning loss to facilitate long-tail learning with extra open-set categories and design a fully automatic data acquisition pipeline to efficiently harvest data from the Web. 
\item We conduct extensive experiments across standard benchmarks using various training paradigms (\eg, random initialization, CLIP~\citep{radford2021clip}, and DINOv2~\citep{oquab2023dinov2}), all of which consistently demonstrate high performance. 

\end{itemize} 

\section{Pilot Study} \label{sec:study}
In this section, we investigate whether granularity impacts performance balance in long-tail distribution. 
We first provide preliminary for long-tail learning and an analysis on a baseline method in \cref{subsec:prelim}. 
Then, we verify the impact of the granularity of training data on long-tail learning (\cref{subsec:granularity_effect}) from both quantitative and qualitative perspectives. 
%



\subsection{Preliminary} \label{subsec:prelim}
In long-tail visual recognition, the model has access to a set of $N$ training samples $\mathcal{S} = \{(\x_n, y_n)\}^N_{n=1}$, where $\x_n \in \mathcal{X} \subset \mathbb{R}^D$ and labels $\mathcal{Y}=\{1,2,..,L\}$. Training class frequencies are defined as $N_y = \sum_{(x_n, y_n) \in\mathcal{S}} \mathds{1}_{y_n=y}$  and the test-class distribution is assumed to be sampled from a uniform distribution over $\mathcal{Y}$
, but is not explicitly provided during training. A classic solution is to minimize the balanced error (BE), of a scorer $\f:\mathcal{X}\rightarrow\mathbb{R}^L$, defined as:
\begin{equation} \label{eq:be}
    {\rm BE}\left(\x,\f(\cdot)\right) = \sum_{y\in\mathcal{Y}}  \prob_{\x|y} \left(y \notin \argmax_{y' \in \mathcal{Y}} \f_{y'}(\x)\right) ,
\end{equation}
where $\f_y(x)$ is the logit produced for true label $y$ for sample $\x$. Traditionally, this is done by minimizing a proxy loss, the Balanced Softmax Cross Entropy (BalCE)~\citep{cui2019class}:
\begin{equation} \label{eq:bal-ce}
    \begin{aligned}
        &\mathcal{L}_{\text{BalCE}}\left(\mathcal{M}(\mathbf{x}|\theta_f,\theta_w), \mathbf{y}_i\right) = - \log\left[p(\mathbf{y}_i|\mathbf{x};\theta_f,\theta_w)\right] \\
        &= - \log\left[\frac{n_{\mathbf{y}_i} e^{z_{\mathbf{y}_i}}}{\sum_{\mathbf{y}_j \in \mathcal{Y}} n_{\mathbf{y}_j} e^{z_{\mathbf{y}_j}}}\right] \\
        &= \log\left[1+\sum_{\mathbf{y}_j\neq \mathbf{y}_i} e^{\log n_{\mathbf{y}_j} - \log n_{\mathbf{y}_i} + \mathbf{z}_{\mathbf{y}_j} - \mathbf{z}_{\mathbf{y}_i}}\right].
    \end{aligned}
\end{equation}
This is known as \textit{re-weighting}, where the contribution of each label's individual loss is scaled by an inverse class frequency derived from the class's instance number $n_{y_i}$. We adopt this setting as the baseline in follow-up experiments.

\noindent \textbf{On the failure of re-balancing.}
The primary challenge of long-tail learning stems from data imbalance, which affects the representation learning of both head classes and few-shot classes. 
For head classes, if there is a lack of effective negative-class samples, then learning an effective boundary is challenging. 
To better demonstrate this, we provide feature visualizations of confusing head (\head{Scottish Deerhound}) and tail (\tail{Irish Wolfhound}) classes on ImageNet-LT in \cref{fig:feature_vis}.
As in \cref{fig:feature_vis_a}, these two classes are challenging even for the advanced vision foundation model DINOv2~\citep{oquab2023dinov2}.
After training with the re-weighting baseline on the imbalanced training data, the learned features seem relatively satisfactory (\cref{fig:feature_vis_b}). However, the generalization is poor: samples in the validation data are still convoluted, and the separation between them is unclear (\cref{fig:feature_vis_c}).
On top of this baseline, we then study the effect of data distribution on long-tail learning. 

\begin{figure}[]
\centering
\includegraphics[width=0.9\linewidth]{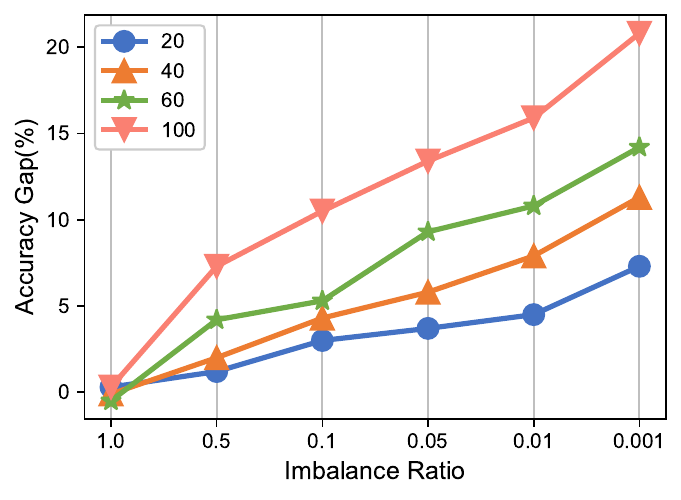}
\vspace{-0.1cm}
\caption{Effect of granularity \vs imbalance ratio.} \label{fig:factor_granularity}
\vspace{-0.4cm}
\end{figure}

\subsection{Granularity Matters in Long-Tail learning} \label{subsec:granularity_effect}


We study whether the granularity of the dataset is critical to long-tail learning. 
Our study is motivated by an intriguing observation that, although more classes and stronger imbalance, we observe nearly balanced performance on iNat18~\citep{van2018inat}, as opposed to ImageNet-LT~\citep{liu2019oltr}. 
A significant distinction is that iNat18 is an extremely fine-grained dataset with over 8000 categories, yet it only consists of 14 superclasses in total. 
On the other hand, ImageNet-LT, although comprising only 1000 categories, has over 100 superclasses, making it relatively coarse-grained.  
Therefore, we conduct experiments to study the effect of granularity on long-tail learning.


\textit{Dataset Configuration.} To this end, we construct a dataset pool using ImageNet-21k~\citep{ridnik2021imagenet21k} and OpenImage~\citep{openimages} datasets. To investigate the influence of granularity, we sample 500 classes from the pool for each time and control the number of superclasses to be $\{20, 40, 60, 100\}$  based on WordNet. 
Then, we used different imbalance ratios $\{1.0, 0.5, 0.1, 0.05, 0.01, 0.001\}$ to study the effect of granularity on the imbalance ratio. 
We train the model (ViT-Base~\citep{dosovitskiy2021vit})  using BalCE~\citep{cui2019class} as \cref{eq:bal-ce}.
We conduct 5 experiments and take the average value.

%
In \cref{fig:factor_granularity}, we show the performance gap between head categories and tail categories under different dataset imbalance ratios.
The results show that as the granularity increases, the dataset is less sensitive to the imbalance ratio. 
For example, when the number of superclasses is 20, the performance gap between the head and tail is 7.3\%, while the gap is 20.8\% when the number of superclasses is 100, under the severe imbalance (imbalance ratio=0.001). 

\begin{mdframed}[backgroundcolor=gray!15] 
\noindent\textbf{Finding 1:}  Increased granularity of training data benefits long-tail learning. 
\end{mdframed}

In a fine-grained long-tail dataset, although there are few samples for tail categories, many categories share similar patterns, which is conducive to learning distinctive features, thus enhancing generalizability. 
As reflected in \cref{fig:feature_vis_d}, for clearer visualization, we sample two fine-grained categories that is denoted as the auxiliary classes. 
The visualization shows that the separation between head and tail classes is clearly improved. 
Also, the distribution of intra-class samples is also more compact. 
Due to the space limitation, we show more examples in Appendix. 
This motivate us to introduce diverse open-set auxiliary categories to enhance the granularity for close-set long-tail learning. 

\begin{mdframed}[backgroundcolor=gray!15] 
\noindent\textbf{Finding 2:} Despite long-tail distribution, increased granularity could explicitly separate and condensify existing data clusters.
\end{mdframed}

Based on the above findings, given a long-tail dataset, we aim to establish a framework that can effectively acquire auxiliary data to enhance the granularity.   
Specifically, we utilize LLMs to query the candidate auxiliary categories and crawl images from the Web, followed by a filtering stage to ensure similarity and diversity. 
To better incorporate auxiliary data for training with target categories, we propose a 
Neighbor-Silencing Loss to avoid being overwhelmed by auxiliary classes. 
Details are included in \cref{sec:method}.

\section{Long-Tail Learning by Category Extrapolation}
\label{sec:method}
In this section, we first introduce our simple and automatic pipeline for obtaining auxiliary data in \cref{subsec:data_searching}. Then, we present our new learning objective that effectively leverages the auxiliary data to enhance long-tail learning in \cref{subsec:training_method}.


\subsection{Neighbor Category Searching} \label{subsec:data_searching}
\begin{figure}[t]
\centering
\includegraphics[width=\linewidth]{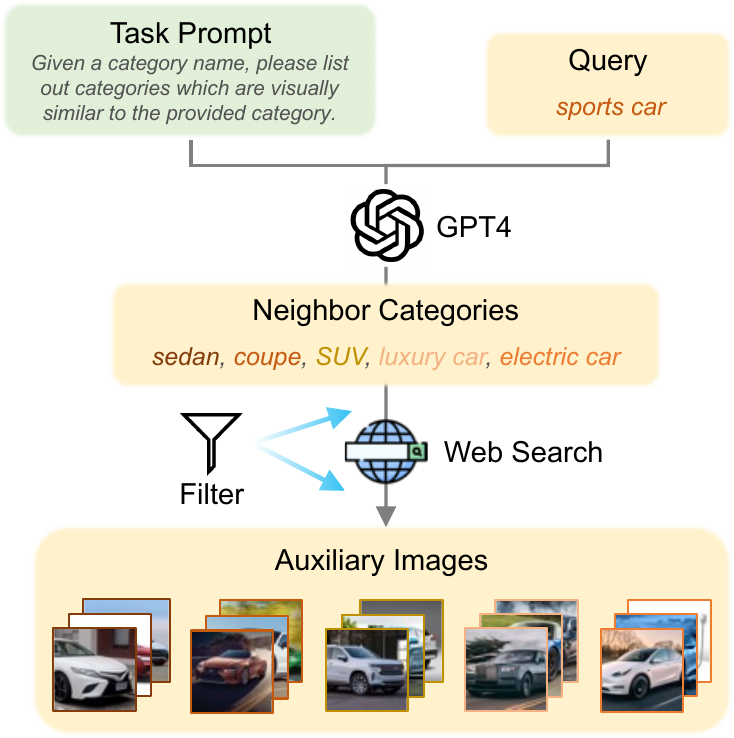}
\caption{{\bf Data crawling pipeline.} We prompt GPT-4~\cite{openai2023gpt4} for fine-grained categories related to query classes and retrieve corresponding images from the web. Classes already in the label set and images of lower visual similarity than the threshold are filtered out.} \label{fig:method-search}
\vspace{-0.3cm}
\end{figure}

In search of neighbor categories sharing some common visual patterns with the pre-defined categories in the dataset, we design a fully automatic crawling pipeline that includes (i) querying neighbor categories from LLMs to obtain similar categories and enhance the granularity of the training data and (ii) retrieving corresponding images from the web and conducting filtering to guarantee similarity and diversity.
An overview of this pipeline is illustrated in \cref{fig:method-search}, and we introduce each step in detail as follows.

\vspace{0.1in}\noindent\textbf{Querying LLM for Neighbor Categories.}
We take advantage of the recent development of Large Language Models (LLMs), \eg, GPT-4~\citep{openai2023gpt4}, and query them for expert knowledge of possible fine-grained classes with respect to the classes to extrapolate (\ie, the medium and tail classes by default).
For example, we can prompt the language model with: ``Please create a list which contains 5 fine-grained categories related to {\tt{\{CLS\}}}''.
However, the output of this naive prompt is unstable, possibly because `fine-grained categories' by itself is quite a broad and vague concept. To make the prompt more concrete and clear for LLMs, we design a structural prompt with in-context learning: 

\noindent

\begin{minipage}{0.45\textwidth}
\vspace{1mm}
\begin{tcolorbox} 
\small
{
\noindent\textbf{Task:} Given a category name, please list out 5 classes that are fine-grained categories related to the provided classes.

\vspace{.25em}

\noindent\textbf{Query:} sports car

\vspace{.25em}

\noindent\textbf{Response:} sedan, coupe, SUV, luxury car, electric car

\vspace{.25em}

\noindent\textbf{Query:} {\tt\{CLS\}}

\vspace{.25em}

\noindent\textbf{Response:}
}
\end{tcolorbox}
\vspace{1mm}
\end{minipage}

The LLM then completes the response above. After that, classes in the target dataset $\mathcal{S}$ are filtered out to avoid possible information leaks.
Then, the remaining class names are fed to an image-searching engine for image retrieval.

\vspace{0.1in}\noindent\textbf{Retrieving  and Filtering Images from the Web.}
Images retrieved by the search engine can be noisy, thus, a filtering strategy is adopted.
An image $\mathbf{x}_r$ corresponding to a specific class $y_i$ is dropped if: (i) the class's name does not exist in the associated caption; or (ii) the visual similarity between the class and this image satisfies thresholds: $  \gamma_1 <  {\rm cos}(\mathbf{p}_i, \mathbf{f}_r) < \gamma_2$.
We employ DINOv2~\citep{oquab2023dinov2} for feature extraction and use cosine similarity as the metric.
Specifically, the prototype $\mathbf{p}_i$ of category $y_i$ is computed as the average feature of all samples of this category in the target dataset $\mathcal{S}$: $\mathbf{p}_i=\nicefrac{1}{n_{y_i}}\sum_j\mathbf{f}_j$.
After the filtering process, the model has access to a set of $M$ auxiliary training samples $\mathcal{A} = \{(\x_m, y_m)\}^M_{m=1}$, where $\x_m \in \mathcal{X} \subset \mathbb{R}^D$ and labels $\mathcal{Y}_a=\{L+1,L+2,..,L+K\}$ and the category number for auxiliary set is $K$.

\subsection{Learning with Auxiliary Categories} \label{subsec:training_method}
We mix the auxiliary dataset $\mathcal{A}$ and the target dataset $\mathcal{S}$ for training.
A naive approach is to directly employ the BalCE loss~\citep{cui2019class} by merging the label space:
\begin{equation} \label{eq:aux-ce}
\begin{aligned}
\mathcal{L}_{\text{BalCE}} = - \log\Biggl[n_{\mathbf{y}_i} e^{z_{\mathbf{y}_i}} / ( \overbrace{\sum_{\mathbf{y}_j \in \mathcal{Y}} n_{\mathbf{y}_j} e^{z_{\mathbf{y}_j}}}^{\text{\footnotesize{Target}}}
+ \overbrace{\sum_{\mathbf{y}_j \in \mathcal{Y}_a} n_{\mathbf{y}_j} e^{z_{\mathbf{y}_j}}}^{\text{\footnotesize{Auxiliary}}})\Biggr].
\end{aligned}
\end{equation}
But note that our objective is to classify $L$ categories within the target dataset, as opposed to $L+K$ categories. 
Directly employing the standard BalCE loss as \cref{eq:aux-ce} would result in an inconsistency between the optimization process and the ultimate goal.
The auxiliary part could overwhelm optimization and result in degenerated performance. We thus ``silent'' them by weighting as follows.

\vspace{0.1in}\noindent\textbf{Silencing the Overwhelming Neighbors.}
%
%
Concretely, if $y_j$ is a neighbor category of $y_i$ from auxiliary categories, we spot this as possible neighbor overwhelming and give the corresponding logit a smaller weight.
To clarify,  $y_j$ is a neighbor category of $y_i$ means that $y_j$ is queried from $y_i$ by Neighbor Category Searching (\cref{subsec:data_searching}).
We thus expect the auxiliary classes to influence less the target class which they are queried from, and contribute more to their classification as a whole with respect to other classes. The neighbor-silencing variant of the re-balancing loss is then formulated as:
\begin{equation} \label{eq:sil-ce}
\mathcal{L}_{\text{NS-CE}}=
\log \left[1+\sum_{\mathbf{y}_j\neq \mathbf{y}_i} \lambda_{ij} \cdot e^{\log n_{\mathbf{y}_j} - \log n_{\mathbf{y}_i} + \mathbf{z}_{\mathbf{y}_j} - \mathbf{z}_{\mathbf{y}_i}}\right],
\end{equation}
where $\lambda_{ij}=\lambda_s$, if $y_i$ and $y_j$ satisfy that one is the other's neighbor category and one of them from auxiliary categories, and $\lambda_{ij}=1$ otherwise. $\lambda_s$ is the weight for balancing the loss between neighbor category pairs and non-pairs. By default, $0<\lambda_s<1$.
In this way, we assign a smaller weight to neighbor category pairs, thus, the effect within neighbor classes is weakened, and the optimization focuses more on their separation as a whole from other confusing classes.
%

\vspace{0.1in}\noindent\textbf{Obtaining the Final Classifier.}
Given that our model's classifier includes more categories, it cannot be directly applied to the target dataset for evaluation. 
A common practice is to discard the trained classifier and re-train it with re-balancing techniques on the target dataset through linear probing~\citep{kang2020decoupling,zhou2020bbn}.
However, this could be suboptimal since the separation hyper-planes shaped by auxiliary categories can be undermined. 
Therefore, we try directly masking out the weights of auxiliary categories, retaining only the weights of the target categories.
Specifically, we denote the trained classifier weights as $\mathcal{W} = \{\mathbf{w}_i\}^{L+K}_{i=1}$, where $\mathbf{w}_i \subset \mathbb{R}^C$, and keep $\mathcal{W} = \{\mathbf{w}_i\}^{L}_{i=1}$.
Surprisingly, this simpler approach works better.
This is potentially because incorporating more auxiliary fine-grained categories can enable the classifier to focus on class-specific discriminative features. 
These features possess stronger generalizability, facilitating the classifier to construct more precise separation hyper-planes.

\section{Experiments} \label{sec:exp}



\subsection{Datasets} \label{subsec:dataset}
We experiment with three standard long-tailed image classification benchmarks.
We report accuracy on three splits of the set of classes: Many-shot (more than 100 images), Medium-shot (20$\sim$100 images), and Few-shot (less than 20 images).
Besides, we also report the commonly used top-1 accuracy over all classes for evaluation.
%

\noindent\textbf{ImageNet-LT}~\citep{liu2019oltr} is a class-imbalanced subset of the popular image classification benchmark ImageNet ILSVRC 2012~\citep{Imagenet}. The images are sampled following the \textit{Pareto} distribution with a power value $\alpha=6$, containing 115.8k images from 1,000 categories.
\noindent\textbf{iNaturalist 2018}~\citep{van2018inat} (iNat18 for short) is a species classification dataset, which consists of 437.5k images from 8,142 fine-grained categories following an extreme long-tail distribution.
\noindent\textbf{Places-LT} is a synthetic long-tail variant of the large-scale scene classification dataset Places~\citep{zhou2017places}. With 62.5k images from 365 categories, its class cardinality ranges from 5 to 4,980.


\begin{table*}[t]
\centering
\renewcommand{\arraystretch}{1.1}
\resizebox{\linewidth}{!}{ 
  \begin{tabular}{llllllllllllll}
  \toprule
  \multicolumn{2}{c}{\multirow{2.5}{*}{\textbf{Method}}}  & \multicolumn{4}{c}{\textbf{ImageNet-LT}} & \multicolumn{4}{c}{\textbf{iNaturalist 18}} & \multicolumn{4}{c}{\textbf{Place-LT}} \\
  \cmidrule(r){3-6} \cmidrule(r){7-10} \cmidrule(r){11-14}
  \multicolumn{2}{c}{}     &\bf Overall &\bf Many &\bf Med. &\bf Few &\bf Overall &\bf Many &\bf Med. &\bf Few &\bf Overall &\bf Many &\bf Med. &\bf Few \\
  \midrule
  \multirow{4}{*}{\rotatebox[origin=c]{90}{Scratch}} & Baseline &60.9 & 72.9 & 56.8 & 41.4  & 76.1 & 78.5 & 76.9 & 74.6 & 39.9 & 43.0 & 40.5 & 33.3 \\
  ~& $+$ RD & 56.8 & 72.1 & 50.4 & 35.8 & 68.4 & 76.4 & 70.1 & 64.3 & 36.5 & 41.9 & 36.1 & 27.5 \\
  ~& $+$ SD &  64.9 & 73.4 & 62.1 & 50.6 & 76.8 & 78.6 & 77.2 & 75.7 & 41.6 & 43.4 & 42.1 & 36.9 \\
  ~ & $+$ {\it Ours} & 68.2$_\text{\bf\textcolor{darkgreen}{$\uparrow$7.3}}$ & 74.5$_\text{\bf\textcolor{darkgreen}{$\uparrow$1.6}}$ & 66.2$_\text{\bf\textcolor{darkgreen}{$\uparrow$9.4}}$ & 57.4$_\text{\bf\textcolor{darkgreen}{$\uparrow$16.0}}$ & 78.0$_\text{\bf\textcolor{darkgreen}{$\uparrow$1.9}}$ & 78.9$_\text{\bf\textcolor{darkgreen}{$\uparrow$0.4}}$ & 78.2$_\text{\bf\textcolor{darkgreen}{$\uparrow$1.3}}$ & 77.5$_\text{\bf\textcolor{darkgreen}{$\uparrow$2.9}}$ & 43.8$_\text{\bf\textcolor{darkgreen}{$\uparrow$3.9}}$ & 43.7$_\text{\bf\textcolor{darkgreen}{$\uparrow$0.7}}$ & 44.8$_\text{\bf\textcolor{darkgreen}{$\uparrow$4.3}}$ & 41.6$_\text{\bf\textcolor{darkgreen}{$\uparrow$8.3}}$ \\
  \midrule
  \multirow{4}{*}{\rotatebox[origin=c]{90}{CLIP}} & Baseline & 74.0 & 77.2 & 72.8 & 68.5 & 75.0 & 77.8 &76.5 &72.5 & 48.4 & 47.9 & 48.6 & 48.9 \\
  ~& $+$ RD & 68.8 & 75.4 & 67.4  & 55.2  & 67.7 & 75.1  & 69.8 & 63.1  & 43.3 & 45.1 &  43.4 & 40.2 \\
  ~& $+$ SD & 75.2 & 77.8 & 74.2 & 71.3 & 76.7 & 78.5 & 77.9 & 74.6 & 49.2 & 48.7 & 49.5 & 49.4 \\
  ~ & $+$ {\it Ours} & 77.3$_\text{\bf\textcolor{darkgreen}{$\uparrow$3.5}}$ & 79.1$_\text{\bf\textcolor{darkgreen}{$\uparrow$1.9}}$ & 76.8$_\text{\bf\textcolor{darkgreen}{$\uparrow$4.0}}$ & 74.1$_\text{\bf\textcolor{darkgreen}{$\uparrow$5.6}}$ & 78.5$_\text{\bf\textcolor{darkgreen}{$\uparrow$3.5}}$ & 79.5$_\text{\bf\textcolor{darkgreen}{$\uparrow$1.5}}$ & 79.3$_\text{\bf\textcolor{darkgreen}{$\uparrow$2.8}}$ & 77.3$_\text{\bf\textcolor{darkgreen}{$\uparrow$4.8}}$ & 50.5$_\text{\bf\textcolor{darkgreen}{$\uparrow$2.1}}$ & 50.0$_\text{\bf\textcolor{darkgreen}{$\uparrow$2.1}}$ & 51.0$_\text{\bf\textcolor{darkgreen}{$\uparrow$2.4}}$ & 50.2$_\text{\bf\textcolor{darkgreen}{$\uparrow$1.3}}$ \\
  \midrule
  \multirow{4}{*}{\rotatebox[origin=c]{90}{DINOv2}} & Baseline & 79.6 & 84.3 & 78.3 & 71.1 & 85.0 & 85.7 & 86.2 & 84.2 & 49.5 & 49.2 & 51.3 & 46.1 \\
  ~& $+$ RD & 77.2 & 83.3 & 75.7 & 65.4 & 75.4 & 82.3 & 76.1 & 72.6  & 45.2 & 47.2 & 45.2 &  41.3\\
  ~& $+$ SD & 80.5 & 83.8 & 79.8 & 73.4  & 85.9 &  85.8 & 86.5 & 85.0 & 49.9 & 49.3 & 51.6 & 47.3 \\
  ~ & $+$ {\it Ours} & 82.0$_\text{\bf\textcolor{darkgreen}{$\uparrow$2.4}}$ & 84.7$_\text{\bf\textcolor{darkgreen}{$\uparrow$0.4}}$ & 81.5$_\text{\bf\textcolor{darkgreen}{$\uparrow$3.2}}$ & 76.2$_\text{\bf\textcolor{darkgreen}{$\uparrow$5.1}}$ & 87.0$_\text{\bf\textcolor{darkgreen}{$\uparrow$2.0}}$ & 86.4$_\text{\bf\textcolor{darkgreen}{$\uparrow$0.7}}$ & 87.4$_\text{\bf\textcolor{darkgreen}{$\uparrow$1.2}}$ & 86.7$_\text{\bf\textcolor{darkgreen}{$\uparrow$2.5}}$ & 50.8$_\text{\bf\textcolor{darkgreen}{$\uparrow$1.3}}$ & 49.4$_\text{\bf\textcolor{darkgreen}{$\uparrow$0.2}}$ & 52.4$_\text{\bf\textcolor{darkgreen}{$\uparrow$1.1}}$ & 49.2$_\text{\bf\textcolor{darkgreen}{$\uparrow$3.1}}$ \\
  \bottomrule
\end{tabular}}
\caption{{\bf Quantitative results of the proposed method on three standard benchmarks.} For each dataset, we conduct three pre-training paradigms (training from scratch, CLIP, and DINOv2) to compare our method with baseline methods on accuracy ($\%$). In addition, we report the \textcolor{darkgreen}{relative improvement} of our method compared to the baseline method in each setting. RD denotes the random auxiliary data and SD is the data from our selected neighbor categories.}\label{tab:main_result} 
\vspace{-0.5cm}
\end{table*}

\subsection{Implementation Details} \label{subsec:implement_details}
We adopt ViT-Base~\citep{dosovitskiy2021vit} as the backbone. 
Our models are trained with the AdamW optimizer~\citep{AdamW} with $\beta_s= \{0.9, 0.95\}$, with an effective batch size of 512.
We train all models with ${\rm RandAug}(9, 0.5)$~\citep{Randaugment}, ${\rm Mixup}(0.8)$~\citep{ref:mixup_iclr2018} and ${\rm Cutmix}(1.0)$~\citep{ref:cutmix_iccv2019}. 
We set the maximum sampling number for each auxiliary category to 50 in each training epoch. 
For the ratio of neighbor category for head, medium, and tail class, we set to $1:\left[\frac{N_h}{N_m}\right]:\left[\frac{N_h}{N_t}\right]$, where $N_h$, $N_m$, and $N_t$ denote the total number of samples of head, medium, and tail classes, respectively. $\left[\cdot\right]$ stands for ceiling, which rounds a number up to the nearest integer.
Following LiVT~\citep{xu2023learning}, the training epochs for ImageNet-LT, iNaturalist, and Place-LT is set to 100, 100, and 30, respectively. 
The hyper-parameter $\lambda_s$ is set to 0.1. 
$\gamma_1$ and $\gamma_2$ are set to 0.7 and 0.98. 
See detailed implementation settings in the Appendix.

\begin{table}[t]
\vspace{0.2cm}
\centering
\centering
\setlength{\tabcolsep}{0.85ex} 
\renewcommand{\arraystretch}{1}
\small
\begin{tabular}{lccccc}
\toprule
\bf Methods &\bf Backbone  &\bf Overall &\bf Many &\bf Med. &\bf Few \\
\midrule
\multicolumn{6}{l}{\bf Training from scratch} \\
\midrule
LiVT~\citep{xu2023learning} & ViT-B  & 60.9 & 73.6 & 56.4 & 41.0 \\
LiVT$^\dagger$~\citep{xu2023learning} & ViT-B   & 65.2 & 73.7 &  62.8 & 49.8  \\
\emph{Ours}  & ViT-B  & \textbf{68.2} & \textbf{74.5} & \textbf{66.2} & \textbf{57.4} \\
\midrule
\multicolumn{6}{l}{\bf Fine-tuning pre-trained model (CLIP)} \\
\midrule
LIFT~\citep{shi2024longtail}  & ViT-B  & 77.0 & 80.2 & 76.1 & 71.5 \\
LIFT$^\dagger$~\citep{shi2024longtail}  & ViT-B  & 77.8 & 80.2 & 77.2 & 73.1 \\
\emph{Ours} & ViT-B  & \bf 78.8 & \bf 80.3 & \textbf{78.4} & \textbf{75.8} \\
\midrule
\multicolumn{6}{l}{\bf Fine-tuning pre-trained model (DINOv2)} \\
\midrule
Bal-CE~\citep{cui2019class} & ViT-B  & 79.6 & 84.3 & 78.3 & 71.1 \\
Bal-CE$^\dagger$~\citep{cui2019class} & ViT-B   & 80.5 & 83.8 & 79.8 & 73.4  \\
\emph{Ours} & ViT-B  & \textbf{82.0} & \textbf{84.7} & \textbf{81.5} & \textbf{76.2} \\
\bottomrule
\end{tabular}
\vspace{0.1cm}
\caption{{\bf Performance on ImageNet-LT.} We report accuracy ($\%$) of all methods under three pre-training paradigms. We also report the performance of adding the auxiliary data but without our method, which denotes by $^\dagger$.} 
\label{table:comp_imagenetlt}
\vspace{-0.6cm}
\end{table}

\subsection{Main Results} \label{subsec:results}

\noindent\textbf{Comparison with Baseline with Different Pre-training.}
We experiment with three different pre-training paradigms (\ie, random initialization, CLIP~\citep{radford2021clip}, and DINOv2~\citep{oquab2023dinov2}). 
The baseline applies Bal-CE~\citep{cui2019class} loss.
As shown in \cref{tab:main_result}, our method significantly improves the performance over the baseline on all three datasets, especially on fewer-shot classes. This improvement is also consistent and generalizes to a variety of pre-training strategies.
In particular, when the model is trained from scratch, we observe a significant performance boost on ImageNet-LT, with a 16.0\% increase in accuracy on the tail classes. 
A plausible explanation is that randomly initialized networks are more prone to overfitting on tail classes compared to large-scale pre-trained models. 
Our method effectively addresses this issue by utilizing neighbor categories.
Besides, even with pre-trained models as initialization, our approach consistently demonstrates satisfactory improvements. 
For example, when using DINOv2 as the backbone, we achieve performance improvements of 5.0\%, 2.5\%, and 3.1\% on the tail classes of ImageNet-LT, iNaturalist, and PlaceLT datasets, respectively, without compromising performance on the head classes.
This verifies our method's generalizability and effectiveness on long-tail datasets. 

\noindent\textbf{Fair comparison.} We also add auxiliary data to the baseline method. 
As shown in \cref{tab:main_result}, RD denotes the random auxiliary data and SD is the data from our selected neighbor categories. 
The results show that, the randomly auxiliary data significantly degrade the performance during the finetuning stage, and the selected neighbor categories can enhance performance. 
Moreover, when using our proposed methods with the neighbor categories, the peformance can be further boosted. 
These results validate the effectiveness of both the auxiliary data and our approach.

\noindent\textbf{Can Learning by Category Extrapolation Enhance the State-of-the-Art Methods?}
We conduct comprehensive experiments with existing SoTAs in \cref{table:comp_imagenetlt}, \cref{table:comp_inat18}, and \cref{table:placeslt}. 
Current methods can be generally categorized into two settings, \ie, training from scratch or adopting CLIP pre-training. 
We also present results obtained by DINOv2, in which we provide the results of Bal-CE~\citep{cui2019class} initialized by pre-trained weights from DINOv2. 
In each pre-training paradigm, we select a SOTA method, and add the same amount of auxiliary data on it, which is denoted by $^\dagger$.
Then we implement our proposed method based on the corresponding SoTA methods. 
%
%
The results show that 1) After adding the auxiliary data from neighbor categories, the performance increase. 2) When using the neighbor categories with our proposed methods, we can further enhance the performance. The potential reason is that our method focuses more effectively on learning the features of target classes, which avoids being overwhelmed by auxiliary categories.
Due to the space limitation, we show more results of previous methods trained with the auxiliary data in Appendix.

\begin{table}[t]
\centering
\setlength{\tabcolsep}{0.85ex} 
\renewcommand{\arraystretch}{1}
\small
\begin{tabular}{lccccc}
\toprule
\bf Method &\bf Backbone  &\bf Overall &\bf Many &\bf Med. &\bf Few \\
\midrule
\multicolumn{6}{l}{\bf Training from scratch} \\
\midrule
LiVT~\citep{xu2023learning} & ViT-B  & 76.1 & 78.9 & 76.5 & 74.8 \\
LiVT$^\dagger$~\citep{xu2023learning} & ViT-B   & 77.0 & 78.8 & 77.4 &  75.9 \\
\emph{Ours}  & ViT-B  & \textbf{78.0} & \textbf{78.9} & \textbf{78.2} & \textbf{77.5} \\
\midrule
\multicolumn{6}{l}{\bf Fine-tuning pre-trained model (CLIP)} \\
\midrule
LIFT~\citep{shi2024longtail}  & ViT-B  & 79.1 & 72.4 & 79.0 & 81.1 \\
LIFT$^\dagger$~\citep{shi2024longtail}  & ViT-B  & 79.5 & 72.9 & 79.4 & 81.3 \\
\emph{Ours} & ViT-B  & \textbf{80.9} & \textbf{79.6} & \textbf{80.1} & \textbf{82.1} \\
\midrule
\multicolumn{6}{l}{\bf Fine-tuning pre-trained model (DINOv2)} \\
\midrule
Bal-CE~\citep{cui2019class} & ViT-B  & 85.0 & 85.7 & 86.2 & 84.2 \\
Bal-CE$^\dagger$~\citep{cui2019class} & ViT-B   & 85.9 &  85.8 & 86.5 & 85.0  \\
\emph{Ours} & ViT-B & \textbf{87.0} & \textbf{86.4} & \textbf{87.4} & \textbf{86.7} \\
\bottomrule
\end{tabular}
\vspace{0.1cm}
\caption{{\bf Performance on iNaturalist 2018.} We report accuracy ($\%$) of all methods under three pre-training paradigms.} 
\label{table:comp_inat18}
\vspace{-0.6cm}
\end{table}

\begin{table}[t]
\centering
\setlength{\tabcolsep}{0.75ex} 
\renewcommand{\arraystretch}{1}
\small
\begin{tabular}{lccccc}
\toprule
\bf Method &\bf Backbone  &\bf Overall &\bf Many &\bf Med. &\bf Few \\
\midrule
\multicolumn{6}{l}{\bf Training from scratch } \\
\midrule
LiVT~\citep{xu2023learning} & ViT-B  & 40.8 & \textbf{48.1} & 40.6 & 27.5 \\
LiVT$^\dagger$~\citep{xu2023learning} & ViT-B  & 42.8 & 48.0 & 42.0 & 35.1 \\
\emph{Ours} & ViT-B  & \textbf{43.8} & 43.7 & \textbf{44.8} & \textbf{41.6} \\
\midrule
\multicolumn{6}{l}{\bf Fine-tuning pre-trained model (CLIP)} \\
\midrule
LIFT~\citep{shi2024longtail}  & ViT-B  & 51.5 & 51.3 & 52.2 & 50.5 \\
LIFT$^\dagger$~\citep{shi2024longtail} & ViT-B  & 51.8 & 51.5  & 52.4 & 51.1 \\
\emph{Ours} & ViT-B   & \bf 52.4 &  \bf 51.6 & \bf 53.0  & \bf 52.3 \\
\midrule
\multicolumn{6}{l}{\bf Fine-tuning pre-trained model (DINOv2)} \\
\midrule
Bal-CE~\citep{cui2019class} & ViT-B & 49.5 & 49.2 & 51.3 & 46.1 \\
Bal-CE$^\dagger$~\citep{cui2019class} & ViT-B  & 49.9 & 49.3 & 51.6 & 47.3\\
\emph{Ours} & ViT-B  & \bf 50.8 & \bf 49.4 & \bf 52.4  & \bf 49.2 \\
\bottomrule
\end{tabular}
\vspace{0.1cm}
\caption{{\bf Performance on Places-LT.} We report accuracy ($\%$) of all methods under three pre-training paradigms.} 
\label{table:placeslt}
\vspace{-0.5cm}
\end{table}

\begin{figure*}[t]
\centering
    \begin{subfigure}[t]{0.3333\textwidth}
        \centering
        \includegraphics[width=\textwidth]{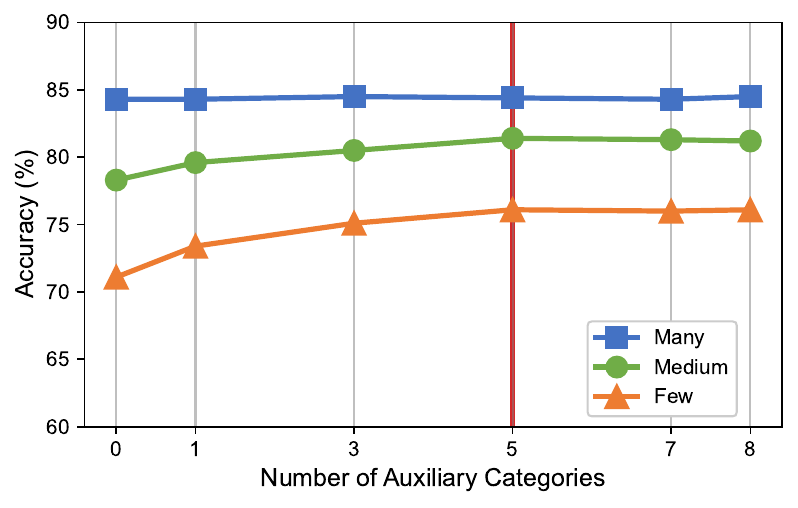}
        \caption{Effect of \#aux. categories.}
        \label{subfig:ablation_cls_num}
    \end{subfigure}\hfill%
    \begin{subfigure}[t]{0.3333\textwidth}
        \centering
        \includegraphics[width=\textwidth]{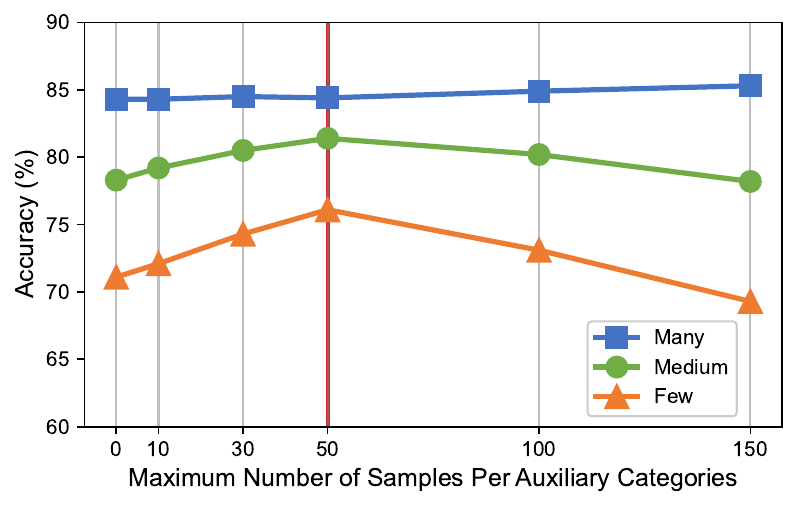}
        \caption{Effect of \#sample/aux. class.}
        \label{subfig:ablation_sample_num}
    \end{subfigure}\hfill%
    \begin{subfigure}[t]{0.3333\textwidth}
        \centering
        \includegraphics[width=\textwidth]{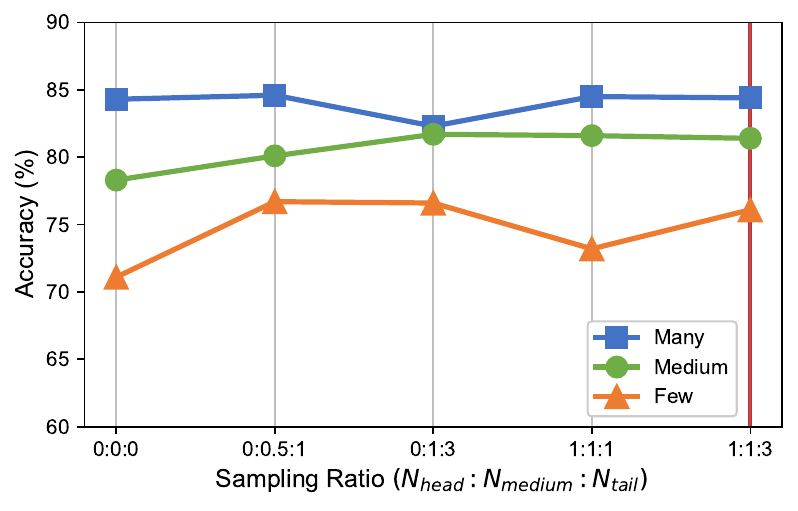}
        \caption{Effect of sampling ratio.}
        \label{subfig:ablation_sample_ratio}
    \end{subfigure}
\vspace{-0.1cm}
   \caption{\textbf{Ablation study on factors related to the curation of auxiliary dataset.} Experiments are conducted on ImageNet-LT~\citep{liu2019oltr}. Default options are marked in \textcolor{BrickRed}{red}.} \label{fig:ablation}
\vspace{-0.35cm}
\end{figure*}

\subsection{Ablation and Analysis} \label{subsec:ablation}

\noindent\textbf{Contributions of Individual Components.}
As shown in Tab. \red{6}, we evaluate the contribution of each component of the full method.
The baseline is BalCE with  DINOv2 pretraining. We conduct ablation experiments on ImageNet-LT. 
%
%
We replace the re-balancing loss (\cref{eq:bal-ce}) with the neighbor-silencing loss (\cref{eq:sil-ce}), obtaining improvements of 1.0\% and 1.9\% in the medium and tail categories, respectively. 
If we use the direct classifier instead of retraining the classfier by linear probing, the performance in the medium and tail categories increases to 79.2\% and 73.2\%, respectively.
The best performance is achieved when we do not re-train the classifier and instead directly utilize the classifier weights corresponding to the target categories.

The curation of the auxiliary dataset primarily involves three hyper-parameters: the number of auxiliary categories associated with a target category, the maximum number of samples per auxiliary class, and the proportion of the number of auxiliary categories for head ($\text{aux}_\text{head}$), medium ($\text{aux}_\text{medium}$), and tail classes ($\text{aux}_\text{tail}$), i.e. $\text{aux}_\text{head}:\text{aux}_\text{medium}:\text{aux}_\text{tail}$ (denoted as auxiliary sampling ratio for simplicity).
We will analyze these three hyper-parameters separately and fix the other two hyper-parameters individually. The default values for these three hyper-parameters are 5, 50, and 1:1:3, respectively.

\noindent\textbf{Number of Sampled Categories.}
\cref{subfig:ablation_cls_num} studies the effect of the number of auxiliary categories for each target class.
The optional values are set to $\{1, 3, 5, 7, 8\}$.
We can observe that as the number of neighbor categories increases, the performance gradually improves and finally saturates when approaching 5.

\noindent\textbf{Maximum Number of Sampled Instances Per Class.}
As shown in \cref{subfig:ablation_sample_num}, we study the effect of the number of samples per neighbor category.
The optional values are $\{10, 30, 50, 100, 150\}$.
If the number of samples collected for a class exceeds the limit, we randomly subsample it to the corresponding number; and if less, we keep them unchanged. 
It can be seen that as the limit increases to 50, the performance improves. 
However, when too many instances are included, the performance drops.
This can be attributed to an excessive number of samples from auxiliary classes, resulting in an overwhelming of these categories.

\begin{table}[t]
\centering
\setlength{\tabcolsep}{1.35ex} 
\renewcommand{\arraystretch}{1}
\small
\begin{tabular}{l|cccc}
\toprule
\bf Methods & \bf Many  & \bf Medium & \bf Few & \bf Overall  \\
\midrule
Baseline & 84.3 & 78.3 & 71.1 & 79.6 \\
+ Random Category & 83.3 & 75.7 & 65.4 & 77.2 \\
+ Neighbor Category & 83.8 & 79.8 & 73.4 & 80.5 \\
+ Neighbor Silencing & 84.3 & 80.8 & 75.3 & 81.4 \\
+ Direct Classifier & \bf 84.7 & \bf 81.5 & \bf 76.2 & \bf 82.0 \\
\bottomrule
\end{tabular}
\vspace{0.3ex}
\caption{{\bf Contributions of individual components.} Results are obtained on ImageNet-LT.} 
\label{tab:ablation}
\vspace{-0.6cm}
\end{table}

\noindent\textbf{Auxiliary Sampling Ratio.}
\cref{subfig:ablation_sample_ratio} studies the proportion of the number of auxiliary categories for head, medium, and tail classes.
When the ratio is 0:1:3, which indicates that the neighbor categories for many classes are removed, we can observe a performance degradation in many classes from 84.4\% to 82.3\%. 
This could be because, with only the addition of auxiliary data in the medium and few-shot categories, feature learning tends to skew towards these medium and few-shot categories.
Moreover, when we decrease the ratio on medium (ratio=1:0.5:3) and tail (ratio=1:1:1) classes, the performance degrades, respectively.

\begin{figure}[t]
\centering
\hspace{-0.2cm}
\includegraphics[width=\linewidth]{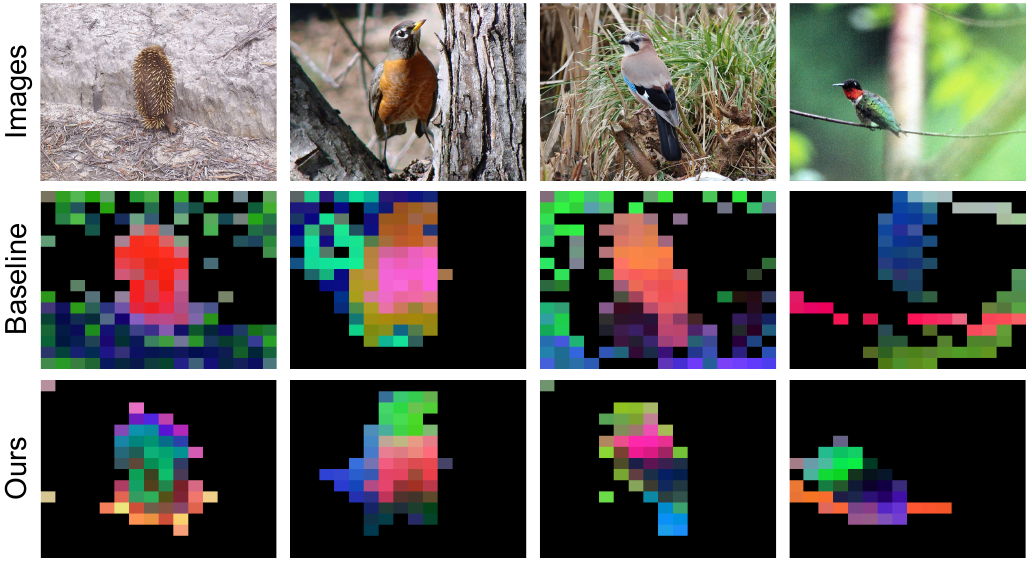}
\caption{{\bf PCA visualization of ``Tail'' images in ImageNet-LT.} Top-3 PCA components of features are mapped to RGB channels, and background is removed by thresholding the first component.} \label{fig:fea-maps}
\vspace{-0.2cm}
\end{figure}

\noindent\textbf{Visualization.}
\cref{fig:fea-maps} shows the top-3 PCA components of images sampled from ``Tail'' classes of ImageNet-LT, where each component is mapped to an RGB channel, and the background is removed by thresholding the first PCA component.
Both the baseline~\citep{cui2019class} and our method adopt DINOv2 pre-training.
While the baseline finds it hard to locate the object of interest, our method clearly captures better objectness despite the scarcity of ``Tail'' images.


\section{Related Works}

\noindent \textbf{Re-Balancing Long-Tail Learning.} \label{subsec:rebalance}
Class-level re-balancing methods include oversampling training samples from tail classes~\citep{chawla2002smote}, under-sampling data points from head classes~\citep{liu2006exploratory}, and re-weighting the loss values or gradients based on label frequencies~\citep{cao2019ldam,cui2019class} or model's predictions~\citep{lin2017focal}.
Classifier re-balancing mechanisms are based on the finding that uniform sampling on the whole dataset during training benefits representation learning but leads to a biased classifier, so they design specific algorithms to adjust the classifier during or after the representation learning phase~\citep{zhou2020bbn,kang2020decoupling}.


\noindent \textbf{Data Augmentation for Long-Tail Learning.} \label{subsec:data_aug}
Spatial augmentation methods have performed satisfactorily for representation learning.
Among these approaches, Cutout~\citep{ref:cutout_2017} removes random regions, CutMix~\citep{ref:cutmix_iccv2019} fills the removed regions with patches from other images, and Mixup series~\citep{ref:mixup_iclr2018,ref:manifoldmixup_2019,summers2019improvedmixup} performs convex combination between images.
Since data augmentation is closely related to oversampling, it is also adopted by recent long-tail recognition literature~\citep{zhou2020bbn,zhong2021mislas}.
These techniques, however, are adopted directly while overlooking special data distributions in long-tail learning.
Recently, Remix~\citep{chou2020remix} was proposed in favor of the minority classes when mixing samples. Yet, this is still bounded by existing classes. 
Unlike above, our method samples images from open-set distributions and could greatly benefit from higher data diversity.


\noindent \textbf{Auxiliary Resources for Long-Tail Learning.} \label{subsec:aux_resource}
Previous efforts mainly lie in refining representations with fixed external image features encoded by pre-trained models~\citep{long2022rac,iscen2023improving}.
The external data could be either the training dataset~\citep{long2022rac} or crawled from the web~\citep{iscen2023improving}, and the fusing process could be either non-parametric~\citep{long2022rac} or learned in an attentive fashion~\citep{iscen2023improving}.
Besides images, another line~\citep{tian2022vl} is to leverage external textual descriptors encoded by vision-language models~\citep{radford2021clip}.
Our method, instead, poses a clear contrast by explicitly introducing external open-set data into a clean training pipeline and is not dependent on any foundation model.
There is also a recent work in self-supervised learning that shares the idea of crawling visually-similar data for task-specific improvements~\citep{li2023internet}. Instead, our work places a special focus on long-tail learning.

\section{Concluding Remarks} \label{sec:conclu}
This paper introduces category extrapolation, which leverages diverse open-set images crawled from the web to enhance closed-set long-tail learning. 
In addition to a clean and decent method that shows superior performance on ``Medium'' and ``Few'' splits across standard benchmarks, we also provide instrumental guidance on when the auxiliary data helps most and empirical explanations on how they help shape the feature manifold through visualizations.
We hope our research will attract more researchers to consider how to leverage additional data to address the pervasive problem in long-tail learning.
Related research topics could include (i) what kind of additional data is more compatible with target datasets and (ii) how to take the additional data in conjunction with target datasets for training.

\clearpage
{
    \small
    \bibliographystyle{ieeenat_fullname}
    \bibliography{main}
}

\appendix

\input{sec/X_suppl}

\clearpage
\setcounter{page}{1}
\maketitlesupplementary

\section{Appendix}

In this supplementary material, we first provide more implementation details in \cref{sec:implementation} about training configurations (\cref{sec:Training}) and auxiliary data collection (\cref{sec:data_collection}). 
Then we conduct additional experiments in \cref{sec:additional_exp} including an experimental comparison to improved SOTA with DIONOv2 (\cref{sec:improved_sota}), and extended ablation studies (\cref{sec:extended_ablation}) related to $\lambda_s$ in the proposed neighbor-silencing loss and the number of samples in the auxiliary dataset, and feature visualization to validate the effectiveness of auxiliary categories (\cref{sec:supp_vis}), and analysis for long-tail in iNaturalist18~\citep{van2018inat} (\cref{sec:extending_tail}). 
In \cref{sec:discussion}, we discuss our contributions (\cref{sec:contributions}), limitations (\cref{sec:limitation}), and future work (\cref{sec:future_work}).


\vspace{-.2cm}

\section{Implementation Details} \label{sec:implementation}

\subsection{Training} \label{sec:Training}

We employ LiVT~\citep{xu2023learning} as our baseline since it achieves the top performance under the training from scratch paradigm using ViT~\citep{VIT}. 
Specifically, when training from scratch, following LiVT~\citep{xu2023learning}, we conduct MAE~\citep{He_2022_CVPR} training on the downstream dataset because training directly on a long-tail dataset with randomly initialized parameters makes it difficult to converge.
When using pre-training paradigms of CLIP and DINOv2, we directly initialize ViT from their weights. 
Furthermore, the models are trained with AdamW optimizer~\citep{AdamW} with $\beta_s= \{0.9, 0.95\}$, with an effective batch size of 512 on 4 NVIDIA 3090 GPUs.
The values for weight decay and layer decay are 0.05 and 0.75, respectively. 
We train all models with ${\rm RandAug}(9, 0.5)$~\citep{Randaugment}, ${\rm Mixup}(0.8)$~\citep{ref:mixup_iclr2018} and ${\rm Cutmix}(1.0)$~\citep{ref:cutmix_iccv2019}. 
Following LiVT~\citep{xu2023learning}, the number of training epochs for ImageNet-LT, iNaturalist 18, and Place-LT is set to 100, 100, and 30, respectively. 
The number of epochs for warmup is set to 10, 10, and 5.
The learning rate is set to 1e-3, 1e-5, and 3.5e-5 for training from scratch, CLIP, and DINOv2, respectively. 
We set a cosine learning rate schedule and the minimum learning rate is 1e-6.
We set the maximum sampling number for each auxiliary category to 50 in each training epoch. 
The hyper-parameter $\lambda_s$ is set to 0.1. 
For the ratio of neighbor category for head, medium, and tail classes, we set to $1:\left[\frac{N_h}{N_m}\right]:\left[\frac{N_h}{N_t}\right]$, where $N_h$, $N_m$, and $N_t$ denote the instance number of head, medium, and tail classes, respectively. $\left[\cdot\right]$ stands for ceiling, which rounds a number up to the nearest integer.

\begin{figure*}[h]
\centering
\includegraphics[width=1.0\linewidth]{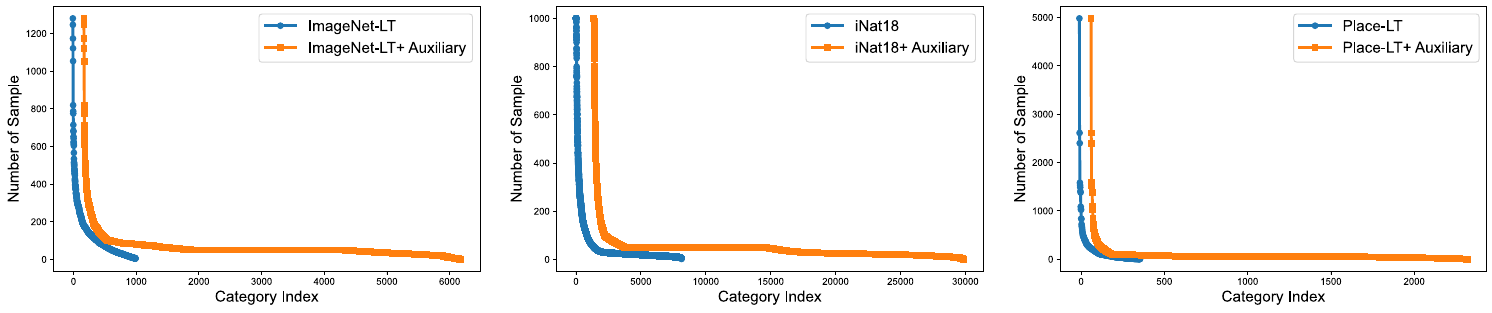}
\caption{{\bf Distribution of samples of original datasets and corresponding datasets with auxiliary data.} Please note that because two lines partially overlap, for a better display, the index of the augmented dataset is slightly shifted. } \label{fig:distribution_info}
  \label{figure:distribution}
  \vspace{0.2cm}
\end{figure*}


\begin{table*}[t]
  \centering
  \scriptsize
  \caption{{\bf Examples of query classes and respective auxiliary classes across three datasets.}}
  \begin{tabular}{p{3.5cm} p{8.5cm}}
    \toprule
    \bf Query & \bf Neighbor Categories \\
    \midrule
    \multicolumn{2}{l}{\bf ImageNet-LT} \\
    \midrule
    Wolf Spider     & Grass Spider, Fishing Spider, Funnel Web Spider, Garden Spider, Dock Spider, huntsman spider\\
    Irish Wolfhound & Greyhound, Pharaoh hound, Silken Windhound, Coonhound, Plott Hound, Bearded Collie\\
    Basketball & Handball, Football, Badminton Shuttlecock, Softball, Cricket Ball, Billiard Ball, Bowling Ball\\
    Kingsnake   & Milk Snake, Corn Snake, Hognose Snake, Ribbon Snak, Black Racer, Speckled Kingsnake\\
    \midrule
    \multicolumn{2}{l}{\bf iNaturalist 18} \\
    \midrule
    Dryopteris Expansa & Dryopteris Austriaca, Dryopteris Carthusiana, Dryopteris Dilatata, Dryopteris Filix-mas\\
    Polypodium Virginianum & Polypodium Amorphum, Polypodium Californicum, Polypodium Vulgare, Polypodium Scouleri\\
    Adiantum Hispidulum & Adiantum Diaphanum, Adiantum Raddianum, Adiantum Reniforme, Adiantum Venustum \\
    Spilosoma Lubricipeda & Arctia Caja, Arctia Villica, Callimorpha Dominula, Diaphora Mendica, Eilema Depressa\\
    \midrule
    \multicolumn{2}{l}{\bf Place-LT} \\
    \midrule
    Bus Interior & Airplane Interior, Tram Interior, Subway Interior, Van Interior, Taxi Interior, Limo Interior\\
    Bamboo Forest & Tropical forest, Evergreen Forest, Pine Forest, Birch Forest, Cypress Forest, Mangrove Forest\\
    Fastfood Restaurant & Seafood Restaurant, Vegetarian Restaurant, Pizza Restaurant, Mexican Restaurant, Steakhouse\\
    Physics Laboratory & Materials Laboratory, Environmental Laboratory, Geology Laboratory, Engineering Laboratory\\
    \bottomrule
  \end{tabular}
  \label{tab:category_list}
\end{table*}

\subsection{Data Collection} \label{sec:data_collection}

We leverage GPT-3.5/4~\citep{openai2023gpt4} to search names of visually similar categories for the downstream long-tail datasets. 
We design a structural prompt with in-context learning and the below shows one example of our interaction with GPT-4~\citep{openai2023gpt4}.
%
\noindent
\begin{minipage}{0.48\textwidth}
\vspace{1mm}
\begin{tcolorbox} 
\small
{
\noindent\textbf{Prompt:} Now I will give you one category name. Please create a list which contains 10 visually similar categories of the provided category.  \\
For example: If I give you a category name: Acacia cochliacantha. 
You should return: [Acacia cambagei, Acacia calamifolia, Acacia campylacantha, Acacia cardiophylla, Acacia colei, Acacia colletioides, Acacia compacta, Acacia corymbosa, Acacia crocophylla, Acacia cuthbertii] \\
Now, I give you this category name: Abaeis Nicippe. \\
You should return:
\vspace{.25em}

\noindent\textbf{Response:} [Eurema ada, Eurema alitha, Eurema andersonii, Eurema beatrix, Eurema blanda, Eurema brigitta, Eurema candida, Eurema celebensis, Eurema desjardinsii, Eurema esakii]}
\end{tcolorbox}
\vspace{1mm}
\end{minipage}

%
\cref{tab:category_list} shows examples of searched category names for each query class on three benchmark datasets. 
The results show that LLM can provide satisfactory responses using our prompts.
After removing duplicates, we obtain 8913, 2318, and 99192 class names for ImageNet-LT~\citep{liu2019oltr}, Place-LT~\citep{zhou2017places}, and iNat18~\citep{van2018inat} datasets, respectively.
Then we search images for each queried name through the web (\textit{e.g.}, Google/Duckduckgo Image Search Engine). 
After removing the dissimilar images, concretely, we collect 4.1M, 1.1M, and 3.6M images in 5012, 1895, and 20380 categories as auxiliary data. 
\cref{figure:distribution} shows the distribution of instance numbers for three datasets in each training epoch. 
It can be observed that 'Tail' is extended by auxiliary data for each dataset.







\begin{table}[h]
\centering
\setlength{\tabcolsep}{.50ex} 
\renewcommand{\arraystretch}{1}
\small
\resizebox{\linewidth}{!}{
\begin{tabular}{l|c|cccc}
\toprule
\bf Methods &\bf Backbone  &\bf Overall &\bf Many &\bf Medium &\bf Few \\
\midrule
\multicolumn{6}{l}{\bf Results on ImageNet-LT with DINOv2 pretraining} \\
\midrule
LiVT(Bal-BCE)~\citep{xu2023learning} & ViT-B   & 79.4 & \textbf{84.9} & 78.2 & 68.5 \\
LiVT(Bal-CE)~\citep{xu2023learning} & ViT-B   & 79.6 & 84.3 & 78.3 & 71.1 \\
\emph{Ours} & ViT-B &    \textbf{81.9} & 84.4 & \textbf{81.4} & \textbf{76.1} \\
\midrule
\multicolumn{6}{l}{\bf Results on iNat18 with DINOv2 pretraining} \\
\midrule
LiVT(Bal-BCE)~\citep{xu2023learning} & ViT-B   & 84.5  & 84.4 & 85.4 & 83.3 \\
LiVT(Bal-CE)~\citep{xu2023learning} & ViT-B   & 85.0 & 85.7 & 86.2 & 84.2 \\
\emph{Ours} & ViT-B &  \textbf{87.0} & \textbf{86.4} & \textbf{87.4} & \textbf{86.7} \\
\midrule
\multicolumn{6}{l}{\bf Results on Place-LT with DINOv2 pretraining} \\
\midrule
LiVT(Bal-BCE)~\citep{xu2023learning} & ViT-B   & 49.6 & \bf 52.4 & 49.7 & 45.2 \\
LiVT(Bal-CE)~\citep{xu2023learning} & ViT-B  & 49.5 & 49.2 & 51.3 & 46.1 \\
\emph{Ours} & ViT-B  & \bf 50.8 &  49.4 & \bf 52.4  & \bf 49.2 \\
\bottomrule
\end{tabular}}
\caption{{\bf Re-implementation of previous method with DINOv2.} We report the performance on three standard benchmark datasets (\textit{i.e.}, ImageNet-LT, iNaturalist 18, and Place-LT). } \label{table:extra_main_results}
\vspace{-0.4cm}
\end{table}



\section{Additional Experiments} 
\label{sec:additional_exp}

\subsection{Comparison to Improved SOTA with DINOv2} \label{sec:improved_sota}
As shown in \cref{table:extra_main_results}, we re-implement LiVT~\citep{xu2023learning} on DINOv2~\citep{oquab2023dinov2}, which is the first work to apply ViT~\citep{VIT} to long-tail learning and leads the performance under the training from scratch paradigm.
Our implementation differs only in that LVIT conducts MAE~\citep{He_2022_CVPR} training on the downstream dataset because training directly on a long-tail dataset with randomly initialized parameters is difficult to converge, whereas we initialize directly with the weight from DINOv2. 
LiVT leverages the Bal-BCE~\citep{xu2023learning} loss by default.
We also implement Bal-CE~\citep{xu2023learning}) to train LiVT with DINOv2. 
\cref{table:extra_main_results} demonstrates that our method shows superior performance on ``Medium'' and ``Few'' splits across three standard benchmarks.
For example, our method surpasses LiVT(Bal-BCE) 3.2\% and 7.6\% on ``Medium'' and ``Few' in ImageNet-LT. 
Note that we set LiVT (Bal-CE) as the baseline method under three pre-training
paradigms (training from scratch, CLIP, and DINOv2).

 \begin{figure*}[t]
\begin{center}
    \begin{subfigure}{0.45\textwidth}
\includegraphics[width=\linewidth]{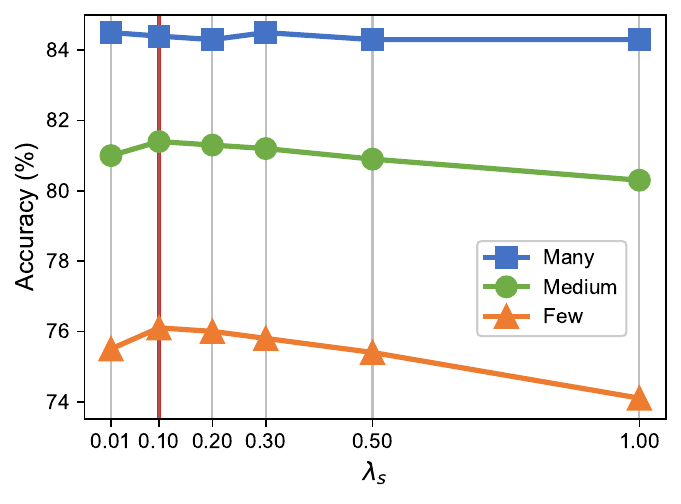}
\caption{\textbf{Ablation study on $\lambda_s$ in the proposed neighbor-silencing loss.} } 
\label{fig:ablation_lamdas}
    \end{subfigure}
    \hfill
    \begin{subfigure}{0.45\textwidth}
\includegraphics[width=\linewidth]{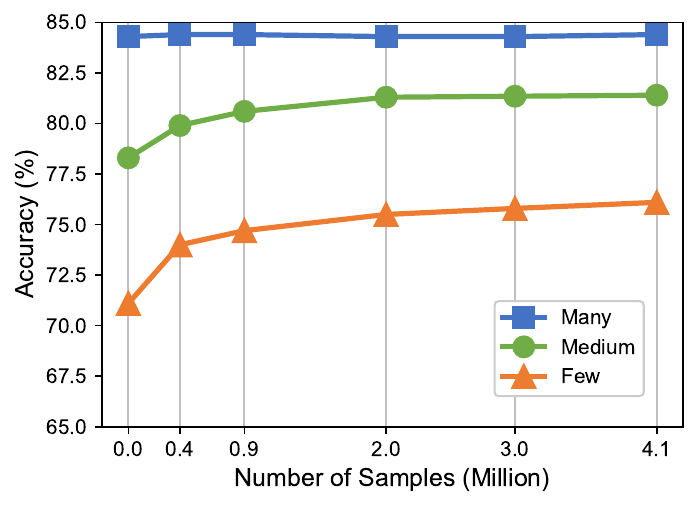}
\caption{\textbf{Ablation study on the number of samples in the auxiliary dataset.} } 
\label{fig:ablation_sample_number}
    \end{subfigure}
    \hfill
   \caption{More ablation studies. Experiments are conducted on ImageNet-LT~\citep{liu2019oltr}.} 
\end{center}
\end{figure*}


\subsection{Extended Ablation Study}
\label{sec:extended_ablation}


\noindent \textbf{Effect of $\lambda_s$}. As shown in \cref{fig:ablation_lamdas}, we study the effect of $\lambda_s$ in the proposed neighbor-silencing loss.
The optional values are $\{0.01, 0.10, 0.20, 0.30, 0.50, 1.00\}$.
It can be seen that as $\lambda_s$ increases to 0.1, the performance improves. 
However, when $\lambda_s$ increases to 1.0, the performance drops.
This can be attributed that as $\lambda_s$ gradually increases, the proposed neighbor-silencing loss will gradually downgrade to the standard cross-entropy loss. 
In this case, the downstream dataset and the auxiliary dataset are treated equally during the training optimization, and the inconsistency between the network's optimization objective and the testing process leads to a decline in performance.

\begin{figure*}
\begin{center}
    \begin{subfigure}{0.85\textwidth}
\includegraphics[width=0.85\linewidth]{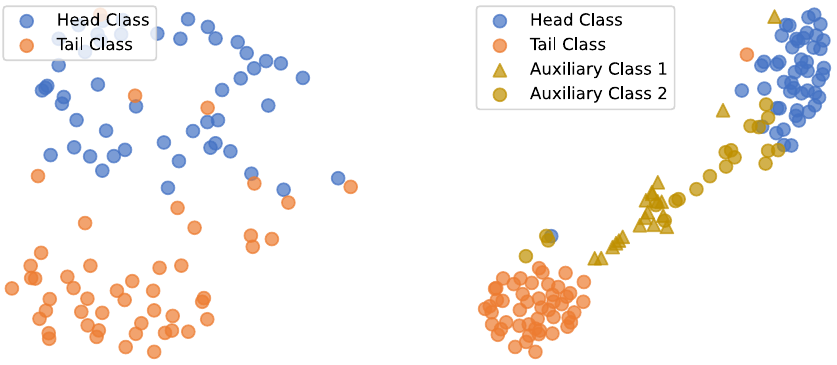}
\vspace{0.2cm}
\caption{Feature visualization of \head{Kit Fox (Head)} and \tail{Cougar (Tail)}.} 
\vspace{0.2cm}
\label{fig:ablation_lamdas}
    \end{subfigure}
    \hfill
    \begin{subfigure}{0.85\textwidth}
\includegraphics[width=0.85\linewidth]{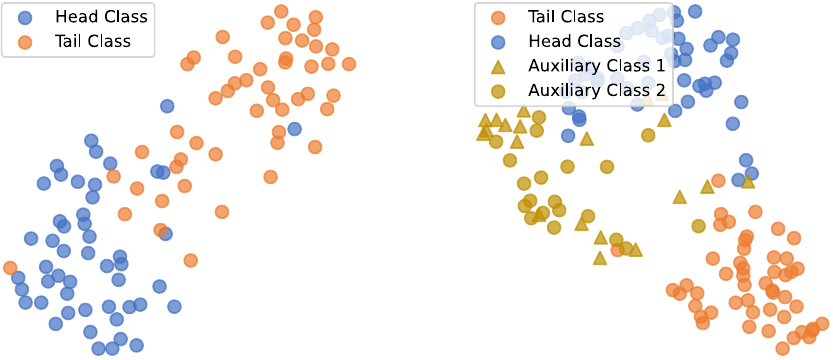}
\vspace{0.2cm}
\caption{Feature visualization of \head{Crane (Head)} and \tail{White Stork (Tail)}.} 
\vspace{0.2cm}
\label{fig:ablation_lamdas}
    \end{subfigure}
    \begin{subfigure}{0.85\textwidth}
\includegraphics[width=0.85\linewidth]{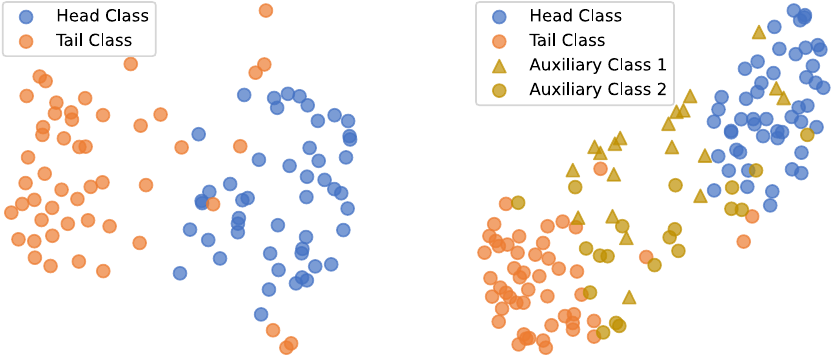}
\vspace{0.2cm}
\caption{Feature visualization of \head{Arctic Fox (Head)} and \tail{Persian Cat (Tail)}.} 
\vspace{0.2cm}
\label{fig:ablation_lamdas}
    \end{subfigure}
    \hfill
    \begin{subfigure}{0.85\textwidth}
\includegraphics[width=0.85\linewidth]{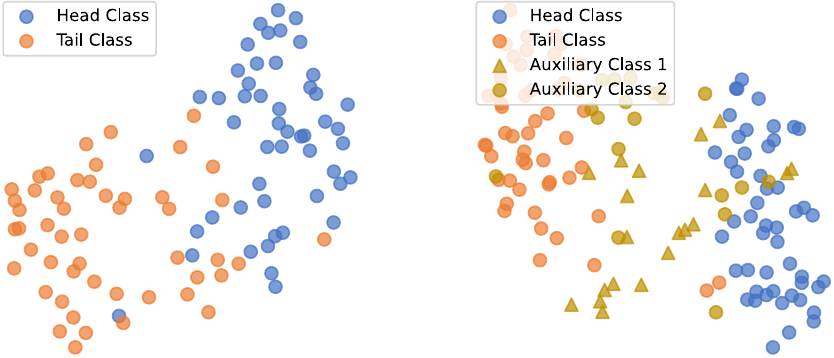}
\vspace{0.2cm}
\caption{Feature visualization of \head{African Hunting Dog (Head)} and \tail{Cheetah (Tail)}.} 
\vspace{0.2cm}
\label{fig:more_vis}
    \end{subfigure}
    \hfill
   \caption{\textbf{Feature visualization of confusing \head{head} and \tail{tail} classes by UMAP~\citep{mcinnes2020umap} on ImageNet-LT~\citep{liu2019oltr}.} The left column shows the feature extracted by the model without auxiliary data, and the right is with the auxiliary fine-grained categories.} \label{fig:supp_vis}
\end{center}
\end{figure*}

\noindent \textbf{Number of Auxiliary Samples}. 
As shown in \cref{fig:ablation_sample_number}, we study the effect of the number of samples in the auxiliary dataset. 
We find that as the number increases from 0 to 0.9 million, there is a dramatic improvement in the accuracy in the few and medium categories, and relatively satisfactory performance is achieved, where +3.7\% and 2.3\% improvement in the few and medium categories, respectively. From 0.9 million to 4.1 million, the performance gradually increases. This indicates the data efficiency of our method.

\subsection{Feature Visualization}\label{sec:supp_vis}
In \cref{fig:supp_vis}, we provide more examples to demonstrate the effectiveness of auxiliary fine-grained categories on the feature separation for the head and tail classes. 
We conduct the experiments on ImageNet-LT~\citep{liu2019oltr} and train the models from random initialization.  
The left column shows the feature extracted by the model without auxiliary data, and the right is with the auxiliary fine-grained categories.
The results show that training with auxiliary fine-grained categories benefits better feature separation between original head and tail classes.

\subsection{Long-Tail in iNaturalist18} \label{sec:extending_tail}


\begin{figure}[]
\centering
\includegraphics[width=0.9\linewidth]{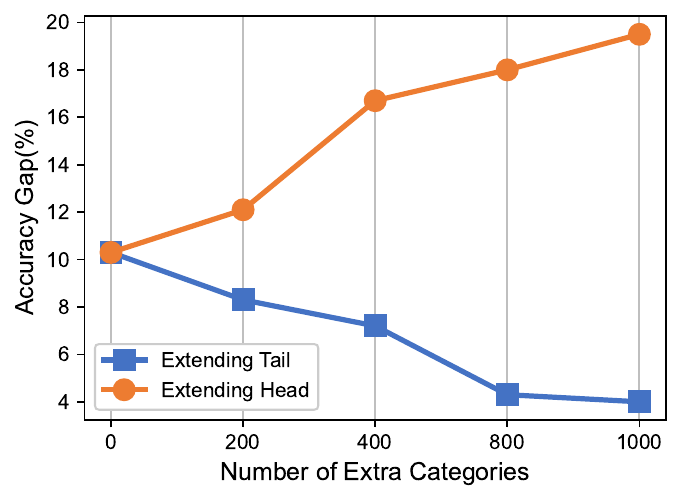}
\vspace{-0.1cm}
\caption{Effect of extending tail vs. extending head.} 
\label{fig:extending_tail}
\vspace{-0.4cm}
\end{figure}

In Sec. \red{3.2}, we validate the effect of granularity on the performance balance. 
Except for the granularity, we find that another difference between iNat18 and ImageNet-LT is that the number of tail categories in iNat18 is significantly larger than the number of head categories.
To validate the effect of the proportion of tail categories, we sample 500 classes from the dataset pool, comprising 60 superclasses, with an imbalance ratio of 0.01.
We conduct two sets of experiments: in the first set, we add extra categories to head classes (each category with more than 100 samples); in the second set, the extra categories are added to tail (each category with less than 20 samples). 
In both sets, the extra categories are fine-grained categories related to the original tail categories. 
As shown in \cref{fig:extending_tail}, the results show that the long-tail benefits the performance balance, while the long-tail will exaggerate the imbalanced performance. 
This also validates our motivation of extending tail categories with fine-grained categories to balance the feature learning.

%



\section{Discussions} \label{sec:discussion}

\subsection{Contributions} \label{sec:contributions}

We summarize and discuss our main contributions as follows: \\
1) \textbf{A new perspective for long-tail learning from neighbor categories.} 
We investigate how to enhance long-tailed learning from open-set data, which is an understudied problem. 
Our pilot study (\textcolor{Red}{Sec. 3}) highlights the granularity matters in long-tail learning (\textcolor{Red}{Sec. 3.2}) and the need for auxiliary categories to improve generalization (\textcolor{Red}{Sec. 3.3}). 
As shown in \textcolor{Red}{Fig. 2(c)}, traditional reweighting methods fail to generalize well. 
However, based on our finding in \textcolor{Red}{Sec. 3.2} that increased granularity of training data benefits long-tail learning ((\textcolor{Red}{Fig. 3})), we apply auxiliary fine-grained categories, which leads to better separation of the target classes (\textcolor{Red}{Fig. 2(d)}).
We also conduct studies on how to select auxiliary categories: inappropriate auxiliary data can even hinder long-tail learning (\textcolor{Red}{Fig. 4}), and there exists a trade-off between the similarity and diversity of auxiliary data (\textcolor{Red}{Sec. 3.3}).
We believe these insights are valuable to the community.
%
\\
\noindent 2) \textbf{Fully automated data acquisition.} Inspired by our findings, we develop a fully automated pipeline for auxiliary data acquisition. 
%
%
As detailed in \textcolor{Red}{Sec. 4.1}, we utilize GPT-4 API to query neighbor categories for target classes. Then, we retrieve images from the Web and automatically filter these images. 
%
%
We will release all the associated code.

\noindent 3) \textbf{A new balanced loss with neighbor silencing.}
As shown in \textcolor{Red}{Sec. 4.2}, we design a new balanced loss with neighbor silencing for improving long-tailed learning with auxiliary data, which mitigates the distraction of extra classes during training.  
After training, we directly mask out the classifier weights of auxiliary categories to obtain the final classifier.
We find that this strategy works better than retraining a new one by linear probing.

\subsection{Limitations} \label{sec:limitation}

This paper proposes to balance feature learning on downstream long-tail datasets by using visually similar categories.
While it has achieved decent performance, there are still the following limitations. 
First, we use LLM~\citep{openai2023gpt4} to obtain the names of similar categories. This step depends on the capability of the large language model; if the model has not seen or is unfamiliar with our query, then this step will fail. 
Second, we obtain images through the web, but we find that some categories are difficult to obtain online, such as those related to the iNat18 categories. 
For some special categories, we may need to look for more specialized websites to crawl data. 
%

\subsection{Future Work} \label{sec:future_work}

In future research, we consider collecting large-scale unlabeled data as an auxiliary dataset for downstream long-tail datasets and then using this dataset to balance feature learning. 
Since it is an unlabeled dataset, we can only consider its similarity to the downstream dataset, so compared to the data collection method in this paper, we can have feature learning on a larger scale. 
Secondly, we find that in a long-tailed distribution dataset, the distribution of superclasses also shows a long-tailed distribution in some datasets (\eg, iNat18~\citep{van2018inat}), we will also take into account the long-tail distribution of superclasses to achieve a better balance in feature learning.


\begin{table}[t]
\vspace{0.2cm}
\centering
\centering
\setlength{\tabcolsep}{0.85ex} 
\renewcommand{\arraystretch}{1}
\small
\begin{tabular}{lccccc}
\toprule
\bf Methods &\bf Backbone  &\bf Overall &\bf Many &\bf Med. &\bf Few \\
\midrule
\multicolumn{6}{l}{\bf Training from scratch} \\
\midrule
BCL~\citep{zhu2022balanced} & ResNet-50  & 56.0 & 67.5 & 52.7 & 34.8 \\
BCL$^\dagger$~\citep{zhu2022balanced} & ResNet-50  & 59.8 & 68.6 & 58.1 & 41.2\\
PaCo~\citep{cui2021paco} & ResNet-50  & 57.0 & 66.4 & 54.5 & 38.6 \\
PaCo$^\dagger$~\citep{cui2021paco} & ResNet-50  & 60.9 & 67.9 & 60.4 & 42.9 \\
NCL~\citep{li2022nested} & ResNet-50  & 57.4 & 67.1 & 54.9 & 38.5  \\
NCL$^\dagger$~\citep{li2022nested} & ResNet-50  & 61.4 & 68.1 & 61.2 & 43.3 \\
\emph{Ours}  & ResNet-50  & 64.5 &  70.1 & 64.8 & 47.9 \\
\midrule
\multicolumn{6}{l}{\bf Fine-tuning pre-trained model (CLIP)} \\
\midrule
BALLAD~\citep{ma2021simple} & ViT-B  & 75.7 & 79.1 & 74.5 & 69.8 \\
BALLAD$^\dagger$~\citep{ma2021simple} & ViT-B & 76.9 & 79.3 & 76.2 & 72.4\\
Decoder~\citep{wang2023exploring}  & ViT-B & 73.2 & 77.9 & 71.9 & 64.7 \\
Decoder$^\dagger$~\citep{wang2023exploring} & ViT-B & 75.2 & 78.1 & 74.9 & 68.6 \\
\emph{Ours} & ViT-B  & \bf 78.8 & \bf 80.3 & \textbf{78.4} & \textbf{75.8} \\
\midrule
\multicolumn{6}{l}{\bf Fine-tuning pre-trained model (DINOv2)} \\
\midrule
Bal-BCE~\citep{xu2023learning} & ViT-B   & 79.7  & 84.1 &  78.5 & 71.3   \\
Bal-BCE$^\dagger$~\citep{xu2023learning} & ViT-B & 80.7 & 84.2 & 80.0 & 73.5\\
\emph{Ours} & ViT-B  & \textbf{82.0} & \textbf{84.7} & \textbf{81.5} & \textbf{76.2} \\
\bottomrule
\end{tabular}
\vspace{0.1cm}
\caption{{\bf Performance on ImageNet-LT.} We report accuracy ($\%$) of all methods under three pre-training paradigms. We also report the performance of adding the auxiliary data, which denotes by $^\dagger$.} 
\label{table:privious_imagenet}
\vspace{-0.6cm}
\end{table}

\begin{table}[t]
\centering
\setlength{\tabcolsep}{0.85ex} 
\renewcommand{\arraystretch}{1}
\small
\begin{tabular}{lccccc}
\toprule
\bf Method &\bf Backbone  &\bf Overall &\bf Many &\bf Med. &\bf Few \\
\midrule
\multicolumn{6}{l}{\bf Training from scratch} \\
\midrule
BCL~\citep{zhu2022balanced} & ResNet-50  & 71.8 & 70.1 & 71.6 & 72.3 \\
BCL$^\dagger$~\citep{zhu2022balanced} & ResNet-50  & 72.9 & 70.3 &  72.9 & 73.4 \\
PaCo~\citep{cui2021paco} & ResNet-50  & 73.2 & 70.4 & 72.8 & 73.6 \\
PaCo$^\dagger$~\citep{cui2021paco} & ResNet-50  & 73.8 & 70.5 & 74.0 & 74.3 \\
NCL~\citep{li2022nested} & ResNet-50 & 74.2 & 72.0 & 74.9 & 73.8 \\
NCL$^\dagger$~\citep{li2022nested} & ResNet-50  & 74.8 & 72.3 & 75.5 & 74.5 \\
\emph{Ours}  & ResNet-50  & \textbf{75.9} & \textbf{74.9} & \textbf{76.2} & \textbf{75.7} \\
\midrule
\multicolumn{6}{l}{\bf Fine-tuning pre-trained model (CLIP)} \\
\midrule
BALLAD~\citep{ma2021simple} & ViT-B  & 75.0  & 77.5 & 75.9 & 73.1 \\
BALLAD$^\dagger$~\citep{ma2021simple} & ViT-B & 77.3 & 78.1 & 77.9 & 76.2 \\
\emph{Ours} & ViT-B  & \textbf{80.9} & \textbf{79.6} & \textbf{80.1} & \textbf{82.1} \\
\midrule
\multicolumn{6}{l}{\bf Fine-tuning pre-trained model (DINOv2)} \\
\midrule
Bal-BCE~\citep{xu2023learning} & ViT-B   & 84.8 & 85.5 & 85.4 &  83.9 \\
Bal-BCE$^\dagger$~\citep{xu2023learning} & ViT-B & 85.6 & 85.9 & 85.8 & 85.1 \\
\emph{Ours} & ViT-B & \textbf{87.0} & \textbf{86.4} & \textbf{87.4} & \textbf{86.7} \\
\bottomrule
\end{tabular}
\vspace{0.1cm}
\caption{{\bf Performance on iNaturalist 2018.} We report accuracy ($\%$) of all methods under three pre-training paradigms. We also report the performance of adding the auxiliary data, which denotes by $^\dagger$.} 
\label{table:privious_inat18}
\vspace{-0.6cm}
\end{table}

\begin{table}[t]
\centering
\setlength{\tabcolsep}{0.75ex} 
\renewcommand{\arraystretch}{1}
\small
\begin{tabular}{lccccc}
\toprule
\bf Method &\bf Backbone  &\bf Overall &\bf Many &\bf Med. &\bf Few \\
\midrule
\multicolumn{6}{l}{\bf Training from scratch } \\
\midrule
PaCo~\citep{cui2021paco} & ResNet-152  & 41.2 & 36.1 & 47.9 & 35.3  \\
PaCo$^\dagger$~\citep{cui2021paco} & ResNet-152  & 42.9 & 37.2 & 48.5 & 40.9 \\
\emph{Ours} & ResNet-152  & \textbf{44.7} & 47.0 &  \textbf{47.1} & \textbf{44.7} \\
\midrule
\multicolumn{6}{l}{\bf Fine-tuning pre-trained model (CLIP)} \\
\midrule
BALLAD~\citep{ma2021simple} & ViT-B  & 49.5 & 49.3 & 50.2 & 48.4 \\
BALLAD$^\dagger$~\citep{ma2021simple} & ViT-B & 50.6 & 50.1 & 51.0 & 50.4\\
Decoder~\citep{wang2023exploring} & ViT-B & 46.8 & 50.6 & 46.8 & 39.6 \\
Decoder$^\dagger$~\citep{wang2023exploring} & ViT-B & 48.8 & 50.8 & 48.4 & 45.8 \\
\emph{Ours} & ViT-B   & \bf 52.4 &  \bf 51.6 & \bf 53.0  & \bf 52.3 \\
\midrule
\multicolumn{6}{l}{\bf Fine-tuning pre-trained model (DINOv2)} \\
\midrule
Bal-BCE~\citep{xu2023learning} & ViT-B & 49.4 & 49.1 & 50.8 & 46.9 \\
Bal-BCE~\citep{xu2023learning} & ViT-B   & 44.9 & 49.3  &  51.5 & 47.5   \\
\emph{Ours} & ViT-B  & \bf 50.8 & \bf 49.4 & \bf 52.4  & \bf 49.2 \\
\bottomrule
\end{tabular}
\vspace{0.1cm}
\caption{{\bf Performance on Places-LT.} We report accuracy ($\%$) of all methods under three pre-training paradigms. We also report the performance of adding the auxiliary data, which denotes by $^\dagger$.} 
\label{table:privious_place}
\vspace{-0.5cm}
\end{table}

\begin{table}[t]
\centering
\setlength{\tabcolsep}{1.35ex} 
\renewcommand{\arraystretch}{1}
\small
\begin{tabular}{l|cccc}
\toprule
\bf Methods & \bf Many  & \bf Medium & \bf Few & \bf Overall  \\
\midrule
Baseline & 84.3 & 78.3 & 71.1 & 79.6 \\
+ OC & 84.4 & 75.4 & 65.9 & 77.6  \\
+ NC &  84.7 &  81.5 & 76.2 &  82.0 \\
+ CD & 84.4 & 83.7  & 83.2  & 83.9  \\
+ CD + NC & 85.4 & 85.0 &  84.8 & 85.1  \\
\bottomrule
\end{tabular}
\vspace{0.3ex}
\caption{{\bf Neighbor Category (NC) v.s. Original Category (OC) from web searching}. Results are obtained on ImageNet-LT. CD denotes the curated data from ImageNet-1k~\cite{Imagenet}.} 
\label{tab:N0_imagenet}
\vspace{-0.6cm}
\end{table}

\begin{table}[t]
\centering
\setlength{\tabcolsep}{1.35ex} 
\renewcommand{\arraystretch}{1}
\small
\begin{tabular}{l|cccc}
\toprule
\bf Methods & \bf Many  & \bf Medium & \bf Few & \bf Overall  \\
\midrule
Baseline & 49.2 & 51.3 & 46.1 & 49.5 \\
+ OC & 49.3 & 46.3  & 40.1 & 46.2  \\
+ NC &  49.4 &  52.4 &  49.2 & 50.8 \\
+ CD & 51.0 & 54.0 & 53.2 & 52.8   \\
+ CD + NC & 52.3 & 55.5 & 54.3 & 54.1   \\
\bottomrule
\end{tabular}
\vspace{0.3ex}
\caption{{\bf Neighbor Category (NC) v.s. Original Category (OC) from web searching}. Results are obtained on Place-LT.  CD denotes the curated data from Places~\cite{zhou2017places}.} 
\label{tab:N0_palce}
\vspace{-0.6cm}
\end{table}

\subsection{Previous Methods on Auxiliary Data}

As shown in ~\cref{table:privious_imagenet}, ~\cref{table:privious_inat18} and ~\cref{table:privious_place}, we conduct experiments on more previous methods. 
We train these methods using the same auxiliary data, which is denoted by $^\dagger$, and conduct the comparison on three pre-training paradigms.  
The results show that the auxiliary data can enhance the performance of previous methods. 
Moreover, our methods can further take advantage of the auxiliary data and  promote the performance. 
The potential reason might be that our method prevents the model from being overwhelmed by auxiliary classes, and ensure alignment with the objectives of the testing phase.

\subsection{Web Searching Data of Original Category} 

We aim to explore whether web searching images of the same categories as the original dataset can benefit long-tail learning.
Using the original category names, we collect corresponding images. 
During training, we ensure the same number of images across experiments. 
As shown in \cref{tab:N0_imagenet} and \cref{tab:N0_palce}, our results indicate that web searching images of the same categories do not improve performance, likely due to a significant distribution gap between the online images and the original dataset.
However, when curated data (CD) is used, we observe a significant performance boost. 
Furthermore, incorporating images from neighboring categories leads to greater improvement.

\clearpage
{
    \small
    \bibliographystyle{ieeenat_fullname}
    \bibliography{main}
}